%% file: main.tex
\documentclass{article}

\usepackage{microtype}
\usepackage{graphicx}
\usepackage{subcaption}
\usepackage{booktabs} 

\usepackage{hyperref}

\usepackage[preprint]{icml2026}

\usepackage{amsmath}
\usepackage{amssymb}
\usepackage{mathtools}
\usepackage{amsthm}

\usepackage[capitalize,noabbrev]{cleveref}

\theoremstyle{plain}

\theoremstyle{definition}

\theoremstyle{remark}

\usepackage[textsize=tiny]{todonotes}
\usepackage{booktabs}
\usepackage{multirow}
\usepackage{colortbl}
\usepackage{xcolor}

\usepackage{tocloft}
\usepackage{etoc}
\usepackage[breakable]{tcolorbox}
\usepackage{xurl}
\usepackage{hyperref}
\usepackage{cleveref}
\usepackage{enumitem}
\usepackage{soul}
\usepackage{cuted}
\usepackage{caption}

\usepackage{etoolbox}

\newcommand{\ACuRL}{\texttt{ACuRL}}
\newcommand{\CUAJudge}{\texttt{CUAJudge}}
\definecolor{diagcolor}{RGB}{225,235,245}

\newcommand{\mean}[1]{$\mathbf{#1}$}
\newcommand{\std}[1]{\scriptsize$\pm#1$}

\icmltitlerunning{Autonomous Continual Learning of Computer-Use Agents for Environment Adaptation}

\begin{document}

\twocolumn[
  \icmltitle{Autonomous Continual Learning \\for Environment Adaptation of Computer-Use Agents}
  \icmlsetsymbol{equal}{*}

  \begin{icmlauthorlist}
    \icmlauthor{Tianci Xue}{osu}
    \icmlauthor{Zeyi Liao}{equal,osu}
    \icmlauthor{Tianneng Shi}{equal,ucb}
    \icmlauthor{Zilu Wang}{osu}
    \icmlauthor{Kai Zhang}{osu}
    \icmlauthor{Dawn Song}{ucb}
    \icmlauthor{Yu Su}{osu}
    \icmlauthor{Huan Sun}{osu}
  \end{icmlauthorlist}

  \icmlaffiliation{osu}{The Ohio State University}
  \icmlaffiliation{ucb}{University of California, Berkeley}

  \icmlcorrespondingauthor{Tianci Xue}{xue.681@osu.edu}
  \icmlcorrespondingauthor{Huan Sun}{sun.397@osu.edu}

  \vspace{-2mm}
  \begin{center}
    \url{https://github.com/OSU-NLP-Group/ACuRL}
  \end{center}
  \vspace{-2mm}
  \icmlkeywords{Machine Learning, ICML}
  \vskip 0.1in
]



\printAffiliationsAndNotice{\icmlEqualContribution}  

\begin{abstract}
Real-world digital environments are highly diverse and dynamic.
These characteristics cause agents to frequently encounter unseen environments and distribution shifts, making continual learning in such environments essential for computer-use agents (CUAs). However, a key challenge lies in obtaining high-quality and environment-grounded training data without relying on costly human annotation.
In this work, we introduce \ACuRL{}, an \textbf{A}utonomous \textbf{Cu}rriculum \textbf{R}einforcement \textbf{L}earning framework that continually adapts agents to specific environments with zero human data. The agent first explores an environment to acquire initial experiences. During subsequent iterative training, a curriculum task generator leverages these experiences together with feedback from the previous iteration to synthesize new tasks tailored for the agent's current capabilities. To provide reliable reward signals, we introduce \CUAJudge{}, a robust automatic evaluator for CUAs that achieves 93\% agreement with human judgments. Empirically, our method effectively enables both intra-environment and cross-environment continual learning, yielding 3–29\% absolute performance gains on the target environments without catastrophic forgetting on others. We also show that it can mitigate performance degradation under environment changes (e.g., version updates, platform migration, and resolution shifts). Further analyses show highly sparse updates (e.g., only 20\% parameters), which helps explain the effective and robust adaptation. 
\end{abstract}

\input{sections/introduction}

\input{sections/methodology}

\input{sections/experiments}

\input{sections/conclusion}

\input{sections/acknowledgments}


\nocite{langley00}

\bibliography{reference}
\bibliographystyle{icml2026}

\input{sections/appendix}


\end{document}

%% file: sections/introduction.tex
\vspace{-6mm}
\section{Introduction}

\begin{figure}[h!]
  \begin{center}
    \centerline{\includegraphics[width=\columnwidth,
      keepaspectratio
      ]
    {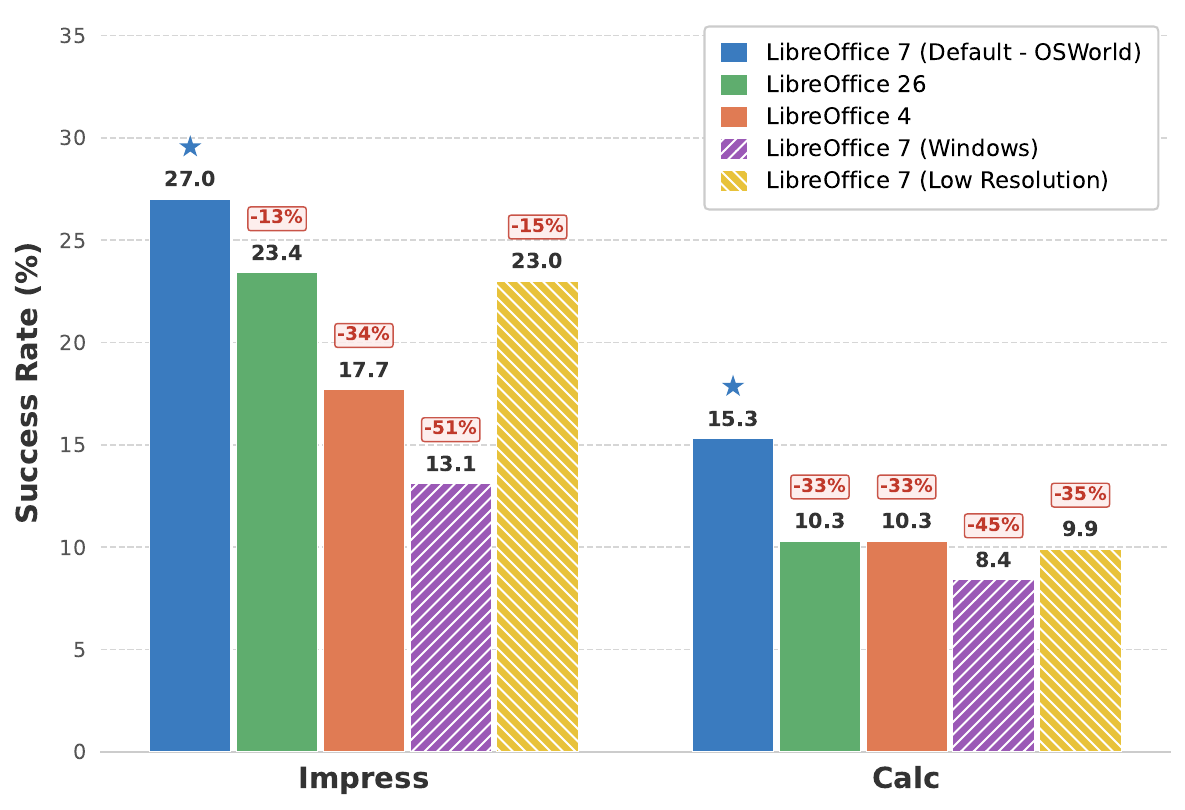}}
    \caption{
      Impact of environmental dynamics, including software updates, platform migration, and resolution changes. For software updates, we evaluate both the {\setlength{\fboxsep}{1pt}\colorbox{green!30}{\strut latest version (LibreOffice 26)}} and a version with {\setlength{\fboxsep}{1pt}\colorbox{orange!30}{\strut major UI changes (LibreOffice 4)}}. For {\setlength{\fboxsep}{1pt}\colorbox{violet!40}{\strut platform migration}}, we evaluate on Windows with the same default version as OSWorld. For {\setlength{\fboxsep}{1pt}\colorbox{yellow}{\strut resolution variation}}, we test a low resolution (1280$\times$800). Here, default OSWorld refers to the benchmark’s default environment configuration on Ubuntu. 
    }
    \label{fig:dynamic_figure}
  \end{center}
  \vspace{-8mm}
\end{figure}

\begin{figure*}[ht!]
  \begin{center}
    \centerline{\includegraphics[width=\textwidth,
      height=0.35\textheight,
      keepaspectratio
      ]
    {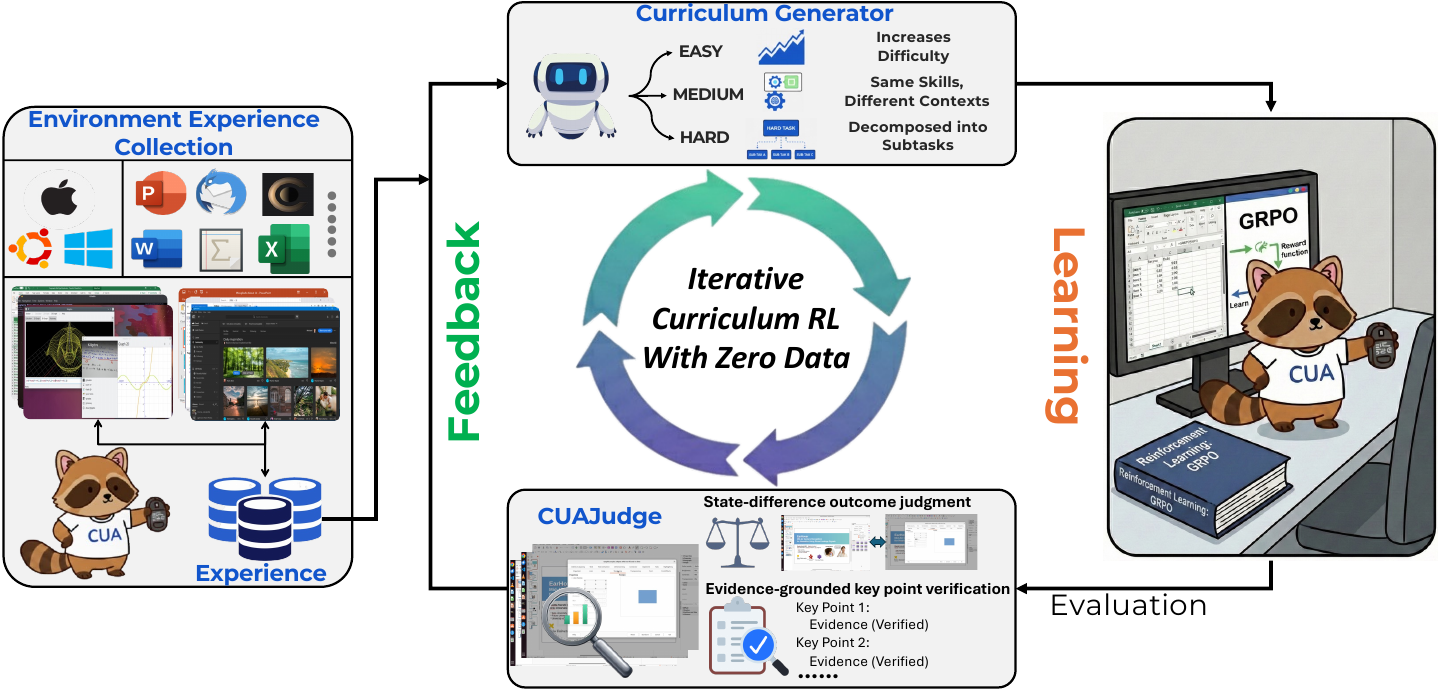}}
    \caption{
      The overview of our \ACuRL{} framework. The agent first autonomously interacts with the environment to collect initial environment experience, and then undergoes iterative RL training to continually learn and adapt to target environments through curriculum tasks with difficulty levels tailored to the agent's current capabilities based on feedback from \CUAJudge{}.
    }
    \label{fig:main_figure}
  \end{center}
\end{figure*}

Driven by rapid advances in Large Language Models (LLMs), autonomous agents capable of interacting with digital worlds, such as browsing the web~\citep{deng2023mind2web,zhou2024webarena,seeact} or using a computer~\citep{xie2024osworld,wu2024copilot,anthropic2024computeruse,wang2025uitars2technicalreportadvancing}, have emerged as a central and highly promising research direction. With active investments, generalist computer-use agents, such as Claude Computer Use~\citep{anthropic2024computeruse,anthropic2025computeruse} and UI-TARS~\citep{qin2025ui,wang2025uitars2technicalreportadvancing}, have shown remarkable improvement on established benchmarks like OSWorld~\citep{xie2024osworld} and Mind2Web~\citep{deng2023mind2web,xue2025an}.

However, strong benchmark performance does not necessarily translate into robustness in real-world applications. First, real-world digital environments are diverse with millions of desktop and web applications, which renders comprehensive coverage during training infeasible. As a result, agents frequently encounter unfamiliar or completely new environments where the functionalities and interaction logic lie beyond their internal knowledge. For example, Claude-3.7's performance drops from 37\% on OSWorld to 10\% on the later released ScienceBoard environments~\citep{sun2025scienceboardevaluatingmultimodalautonomous}. 
Second, real-world digital environments are inherently dynamic~\citep{pan2024webcanvas,xue2025an}: they evolve continuously, with underlying functionality and user interface being updated over time. Such changes cause distribution shifts for agents trained under static assumptions~\citep{ishmam2026timewarp}. As shown in Figure~\ref{fig:dynamic_figure}, variations such as software updates, platform migration, and resolution changes consistently degrade performance, with relative drops of up to 51\%.
These challenges highlight \textit{continual learning in specific target environments} as a central requirement for real-world applicability, which remains unexplored.

In this work, we make the first effort to fill the gap and introduce \ACuRL{} (Figure~\ref{fig:main_figure}), an autonomous curriculum reinforcement learning framework that effectively enables agents to continually learn in target environments without any human supervision. The framework begins with the agent autonomously exploring target environments under diverse web-crawled contexts to gather initial environment experiences, including contents, interfaces, and functionalities. The agent is then trained through an iterative reinforcement learning (RL) to progressively learn environment-related knowledge. At the end of each iteration, an automatic evaluator will assess the current agent’s capabilities, and a curriculum generator uses this feedback alongside accumulated experiences to synthesize a new set of tasks. The difficulty level of the new tasks is tailored to the agent’s current capability. 
To provide reliable evaluation reward signals, we further introduce \CUAJudge{}, an automatic evaluator for CUAs, which achieves 93\% agreement with human judgments. 
\CUAJudge{} performs a difference analysis between initial and final environment states and verifies each key point for task completion based on supporting screenshots and actions, enabling reliable and interpretable outcome assessment. 
Finally, to address the challenges of scaling RL to large-scale, realistic, and dynamic digital environments, we introduce a lightweight unified environment management protocol that supports batched environment creation and deletion with fault tolerance. Combined with asynchronous environment preloading and evaluation, our design significantly accelerates the training process.

Empirical results across six representative environments demonstrate the effectiveness of our approach for both intra-environment and cross-environment continual learning scenarios, yielding absolute performance improvements of 3–29\% across iterations on target environments without catastrophic forgetting in others. 
Furthermore, we evaluate \ACuRL{} under realistic environment changes in Figure~\ref{fig:dynamic_acurl} and show that it effectively mitigates their impact, achieving relative performance gains of up to 145\% under such shifts. 
To better understand how the agent evolves during continual learning, we conduct a detailed analysis of the agent’s internal parameter updates. We find that only about 20\% of the parameters are substantially updated, consistent with recent findings on RL~\citep{mukherjee2025reinforcement}. Moreover, the update patterns differ markedly between the LLM backbone and the vision encoder. Specifically, parameter updates in the LLM backbone are primarily concentrated in the higher layers, whereas updates in the vision encoder gradually diminish as depth increases. This analysis helps explain why the agent can continually learn in target environments while preserving performance in other environments, thereby mitigating catastrophic forgetting.

%% file: sections/methodology.tex
\section{Continual Learning of CUAs and Related Work}

Real-world digital environments are diverse and evolving; as a result, agents frequently encounter unfamiliar or completely new
environments where little or zero data are seen during training.
Prior work relies on fine-tuning with human-annotated, domain-specific data, which is costly, unscalable, and generalizes poorly to unseen environments~\citep{wang2025opencuaopenfoundationscomputeruse,liu2025scalecua,he2025efficient}. Moreover, both environment dynamics (e.g., software updates and platform migration) and user needs (e.g., new environments and more complex goals) evolve sequentially over time, rendering the static-data assumption inadequate and making continual learning essential.
Building on this, we formulate the problem as \emph{continual learning of computer-use agents}, where a base agent with agentic capabilities (e.g., UI grounding and planning) continually learns from a sequential set of tasks $\mathcal{U} = (T^{(1)}, T^{(2)}, \ldots, T^{(N)})$ over time. 
We consider two continual learning scenarios~\citep{zheng2026lifelong} that differ in the scope of the environment: (i) \textbf{\emph{intra-environment continual learning}}, where tasks come from a single environment with shared interface and dynamics, but exhibit different distributions across different iterations, where they progressively vary in goals, required skills, and complexity and require the agent to continually expand its capabilities over time. 
(ii) \textbf{\emph{cross-environment continual learning}}, where the agent \emph{sequentially} learns from multiple heterogeneous environments with distinct interfaces and dynamics. This requires the agent to continually adapt and transfer knowledge across environments while maintaining performance on previously learned ones without catastrophic forgetting.
Our formulation of continual learning for CUAs is consistent with the general continual learning literature~\citep{legg2008machine,zheng2026lifelong}, where an agent incrementally learns from a sequence of tasks or environments, adapts to evolving distributions, and retains previously acquired knowledge without catastrophic forgetting. Additional discussion of related work is provided in Appendix~\ref{appendix:related_work}.

\noindent \textbf{Challenges.}
One fundamental challenge in continual learning of CUAs lies in acquiring high-quality, environment-grounded agent data in new environments.
While traditional continual learning methods such as Elastic Weight Consolidation~\citep{kirkpatrick2017overcoming}, PackNet~\citep{mallya2018packnet}, and Progressive Networks~\citep{rusu2016progressive} have been explored, they primarily focus on mitigating catastrophic forgetting and assume access to high-quality training data, without addressing how such data can be acquired in complex, real-world environments.
Recent methods based on synthetic tasks and RL~\citep{kuba2025language,huang2025r,liu2025spiralselfplayzerosumgames} break free from human effort and have shown promise for math and coding tasks. However, they are not directly applicable to CUAs due to two key challenges: (1) \textbf{\emph{Task generation.}} Computer-use tasks must be grounded in dynamic, multimodal environmental contexts (e.g., different initial states of software) that are rarely present in pretraining data, making environment-agnostic generation unreliable~\citep{kuba2025language,zhao2025absolute}. Moreover, as the agent improves, generating tasks that match its evolving competence remains an additional challenge for effective continual learning.
(2) \textbf{\emph{Trajectory evaluation.}} Unlike domains with verifiable rewards readily available such as math and coding~\citep{zhao2025absolute,guan2025rstarmath}, computer-use task completion cannot be reliably determined from the final state alone but needs to refer to a long trajectory of multimodal observations and actions that often introduces the long context issue, making it difficult to obtain reliable reward signals during RL. 
\section{ACuRL}
To address the aforementioned challenges, we introduce \ACuRL{}, an \textbf{A}utonomous \textbf{Cu}rriculum \textbf{R}einforcement \textbf{L}earning framework.
Specifically, for the \textit{task generation} challenge, the agent first interacts with target environments under diverse contexts to accumulate environment-specific experience, thereby grounding task generation in the target environmental contexts {(\S\ref{method:cua_task_generation}). To support effective continual learning, we further design a curriculum task generation scheme (\S\ref{method:iterative_rl_curriculum_learning}) that synthesizes tasks tailored to the agent’s evolving capabilities. For the \textit{trajectory evaluation} challenge, we introduce \CUAJudge{} {(\S\ref{method:automatic_evaluator}) to accurately evaluate task completion over long-horizon trajectories.

\subsection{Computer-Use Task Generation}
\label{method:cua_task_generation}
To ground generated CUA tasks in dynamic, multimodal environmental contexts, we introduce \textit{environment exploration} and \textit{context review} stages to acquire some environment experience, ensuring that generated tasks are realistic, executable, and grounded in diverse environmental contexts.

\noindent \textbf{Environment Exploration.}
This stage aims to collect environment-specific experience for the task generator $G$ within a target environment $\mathcal{E}$, including its interface and functionalities, so that it can synthesize high-quality and valid tasks. We let the agent $M$ autonomously explore the environment $\mathcal{E}$ without any constraints, interacting with diverse interface elements and functions. The resulting action sequences $A = (a_1, a_2, \ldots, a_i)$ and observations $O = (o_1, o_2, \ldots, o_i)$ are recorded as \emph{exploration trajectories}
$\tau^{\text{exp}} = (o_1, a_1, o_2, a_2, \ldots, o_i, a_i)$,
which serve as preliminary experience for subsequent task generation by $G$.

\noindent \textbf{Context Review.}
For each target environment $\mathcal{E}$, we first collect diverse real-world contexts $C=(c_1, c_2, \ldots, c_i)$ via web crawling, where each context corresponds to different user-created artifacts 
(e.g., documents, emails or slides; examples are provided in Appendix~\ref{appendix:example_contexts}).
The agent $M$ then actively interacts with the environment to review these contexts (e.g., navigating through each slide).
The resulting interaction sequences are recorded as \emph{context trajectories} $\tau^{\text{ctx}}$. Conditioning the task generator $G$ on these observed context trajectories significantly increases task diversity and better captures the complexity of real-world user requests. See Appendix~\ref{app:context_review} for more details.

\subsection{Iterative RL with Curriculum Learning}
\label{method:iterative_rl_curriculum_learning}
Traditional RL paradigms typically rely on a fixed task pool or filtering strategies to select informative ones~\citep{yu2025dapo,zheng2025act,Polaris2025}, without adapting to the agent’s evolving capabilities. Consequently, overly difficult tasks remain unsolved due to limited intermediate supervision, while trivial tasks continue to be sampled instead of being replaced by more challenging ones, hindering effective continual learning~\citep{ladosz2022exploration,diaz-bone2025discover}.
This issue is particularly pronounced in the CUA domain, where agents usually perform poorly in target environments. To address this, we introduce an iterative curriculum RL framework that \textit{dynamically adjusts task difficulty during the agent’s learning progress}. Specifically, the agent is trained for $N$ iterations, each with $x$ optimization steps. 
After each iteration, we perform \textit{capability evaluation} on the current tasks, and use the feedback to guide $G$ in generating curriculum tasks with adjusted difficulty for the next iteration.

\noindent \textbf{Capability Evaluation.}
At the end of iteration $n$, the agent is evaluated on task set $T^{(n)}$. For each task $t_k^{(n)} \in {T}^{(n)}$, we conduct $m$ rollouts and quantify performance as the mean success rate across these rollouts. $y_{k,j}^{(n)}$ denotes whether the agent successfully completes task $t_k^{(n)}$ in the $j$-th rollout.

\begin{equation}
s_k^{(n)} = \frac{1}{m} \sum_{j=1}^{m} \mathbb{I}\!\left( y_{k,j}^{(n)} = 1 \right),
\end{equation}

\noindent \textbf{Curriculum Task Generation.}
Drawing on the feedback from the capability evaluation at iteration $n$, we generate a new set of tasks $T^{(n+1)}$ with different skills tailored to the agent's current capability.
Here, we use the term skill to refer to a recurring, reusable, and semantically meaningful sub-goal or interaction pattern required to complete a task.
Formally, given the previous task set $T^{(n)}$ and evaluation feedback $\{s_k^{(n)}\}$, context trajectories $\tau^{\text{ctx}}$, exploration trajectories $\tau^{\text{exp}}$, the generator produces a new task set for the next training iteration
$T^{(n+1)} \sim G(\cdot \mid \tau^{\text{ctx}}, \tau^{\text{exp}}, T^{(n)}, \{s_k^{(n)}\})$.
Specifically, tasks in $T^{(n)}$ are partitioned into three difficulty categories based on thresholds $\delta_{\text{high}}$ and $\delta_{\text{low}}$ to guide the adaptive task generation process: (i) \textbf{Easy Level ($s_k^{(i)} > \delta_{\text{high}}$):} As the agent masters the core skills required for these tasks, the generator will synthesize more challenging variants by increasing complexity, such as incorporating additional skill requirements or extending the task horizon. (ii) \textbf{Medium Level ($\delta_{\text{low}} \le s_k^{(i)} \le \delta_{\text{high}}$):} Since the agent has not yet fully mastered the required skills, the generator will promote diversity while preserving learnability by synthesizing tasks that share core skills but vary in scenarios or personas.
(iii) \textbf{Hard Level ($s_k^{(i)} < \delta_{\text{low}}$):} For tasks that remain difficult, the generator applies hierarchical decomposition, breaking them into meaningful subtasks to isolate essential skills and enable intermediate learning. 
Specifically, the generator first identifies underlying sub-steps and required skills (e.g., invoking functions or manipulating objects), then creates simpler, skill-focused but realistic tasks that are grounded in user-created artifacts. 

Through this progressive task generation, the curriculum generator dynamically adjusts task difficulty with the agent’s evolving capabilities, enabling effective continual learning. See Appendix~\ref{appendix:details_curriculum_task_generation} for details.
To assess the validity of the generated tasks, we conduct a human evaluation on 144 tasks, uniformly sampled across iterations and six environments. Results show that 94\% are valid (i.e., meaningful and executable), demonstrating the effectiveness of our method. Beyond validity, we further analyze how the synthesized tasks evolve across iterations and under different distributions. As shown in Figure~\ref{fig:task_complexity}, both average task length and the number of agent execution steps increase steadily, serving as a proxy for growing task complexity. 
To further corroborate this trend, we qualitatively examine representative examples.
Additional details are provided in Appendix~\ref{appendix:task_validity} with examples in Table~\ref{tab:task_example}. Together, these results indicate that our method co-evolves learning signals with the agent’s capabilities, enabling effective continual learning for CUAs.
\subsection{Automatic Evaluator}
\label{method:automatic_evaluator}
Another fundamental challenge in computer-use agent learning lies in obtaining reliable reward signals. 
To this end, we propose \CUAJudge{}, a reliable automatic evaluator for CUA trajectories. Our design builds on the WebJudge~\citep{xue2025an} pipeline with three stages: key point identification, key screenshot identification, and outcome judgment. Specifically, key point identification extracts task-critical requirements from the task description; key screenshot identification selects informative screenshots from the trajectory; and outcome judgment determines task completion by reasoning over the task description, key points, key screenshots, and action history. However, WebJudge is primarily tailored to web tasks, which can be viewed as a subset of the broader computer-use tasks. In general, computer-use tasks are more complex, requiring fine-grained examination of internal system and software states~\citep{gonzalez2025unreasonable,agasheagent}, whereas web tasks often only involve verifying surface-level effects, such as whether a specific filter has been applied.
Accordingly, \CUAJudge{} extends WebJudge with two key modifications to address the unique challenges of general computer-use tasks and improve evaluation reliability:
\begin{itemize}[leftmargin=*, topsep=0pt, itemsep=0pt, parsep=0pt]
    \item \textbf{State difference analysis.}
    Many computer-use tasks (e.g., editing slides or generating plots) can be reliably verified by comparing differences between the initial and final environment states. Accordingly, we augment outcome judgment with explicit state-difference analysis that highlights these changes and incorporates them into the final decision. ~\citet{gonzalez2025unreasonable} also introduce a state-difference module based on step-wise screenshot changes to reduce noise in long trajectories. In contrast, we directly compare initial and final states to determine whether the final state satisfies task requirements. Moreover, their method is used to select a trajectory from a candidate set, whereas \CUAJudge{} assigns a binary reward to each individual trajectory.
    \item \textbf{Evidence-grounded key point verification.}
    \CUAJudge{} explicitly evaluates the completion of each key point and requires concrete evidence for every decision, including the supporting screenshots and actions. This evidence-grounded design substantially improves evaluation precision.
\end{itemize}
\subsection{Optimization}
Following~\citet{wei2025webagent}, we extend the Group Relative Policy Optimization (GRPO) algorithm~\citep{shao2024deepseekmath} to a multi-turn setting by decomposing each trajectory into a sequence of step-level action groups and optimizing them jointly.
For a given task $I$, we sample a group of $G$ trajectories in the environment
$\{\tau^{(i)}\}_{i=1}^{G}$. 
Each trajectory
$\tau^{(i)} = \{(s_t^{(i)}, a_t^{(i)})\}_{t=1}^{T^{(i)}}$
consists of multiple interaction steps, where $s_t^{(i)}$ denotes the observation at step $t$
and $a_t^{(i)}$ denotes an action generated autoregressively.
We optimize the policy by minimizing the objective:
\begin{equation}
\mathcal{L}(\theta)
=
-\frac{1}{G}
\sum_{i=1}^{G}
\frac{1}{T^{(i)}}
\sum_{t=1}^{T^{(i)}}
\frac{1}{|a_t^{(i)}|}
\sum_{k=1}^{|a_t^{(i)}|}
\Big[
\tilde{A}^{(i)}_{t,k}
-
\beta\, D_{\mathrm{KL}}(\theta)
\Big],
\end{equation}

where $\tilde{A}^{(i)}_{t,k} = \min\!\big\{ r^{(i)}_{t,k}(\theta)\,\hat{A}^{(i)},
\mathrm{clip}\big(r^{(i)}_{t,k}(\theta), 1-\epsilon, 1+\epsilon\big)\,\hat{A}^{(i)} \big\}$
is the clipped surrogate objective term for the $k$-th token in the action $a_t^{(i)}$ of trajectory
$\tau^{(i)}$ and the importance sampling ratio for the $k$-th token of the action at step $t$ is
defined as 
$r^{(i)}_{t,k}(\theta)
=
\frac{
\pi_\theta(a^{(i)}_{t,k} \mid I, s^{(i)}_t, a^{(i)}_{t,<k})
}{
\pi_{\theta_{\mathrm{old}}}(a^{(i)}_{t,k} \mid I, s^{(i)}_t, a^{(i)}_{t,<k})
}.
$
Following GRPO, we compute a trajectory-level normalized advantage using the group of sampled
trajectories: $
\hat{A}^{(i)} =
\frac{R^{(i)} - \mathrm{mean}(\{R^{(j)}\}_{j=1}^{G})}
{\mathrm{std}(\{R^{(j)}\}_{j=1}^{G})},
$
where $R^{(i)}$ denotes the reward for trajectory $\tau^{(i)}$ produced by \CUAJudge{}.
We assign the same trajectory-level advantage $\hat{A}^{(i)}$ to all actions within the trajectory.

%% file: sections/experiments.tex
\section{Experiments}
\label{experiments}
\subsection{Experiment Setup}
\noindent \textbf{Benchmarks and Metrics.}
We consider six representative environments: LibreOffice Impress, LibreOffice Writer, and LibreOffice Calc from OfficeWorld~\citep{lai2025computerrl} and OSWorld~\citep{xie2024osworld} for office productivity tools (we refer to them as Impress, Writer, and Calc for simplicity); Thunderbird from OSWorld for email management; and Celestia and KAlgebra from ScienceBoard~\citep{sun2025scienceboardevaluatingmultimodalautonomous} as specialized scientific software.
Office tools represent some of the most commonly used software, where users expect agents to adapt effectively to support everyday productivity tasks. Thunderbird, while also a real-world email client, is less prevalent in pretraining data and thus presents a less familiar environment. In contrast, Celestia and KAlgebra are professional scientific applications for astronomical visualization and symbolic mathematics, which are niche and serve as testbeds for continual learning and adaptation to new environments.
For evaluation, we use tasks from OSWorld, OfficeWorld, and ScienceBoard, and report success rates based on each benchmark’s rule-based protocol.
Table~\ref{tab:task_statistics} in Appendix~\ref{appendix:experimental_details} presents the task statistics for each environment.
To mitigate the effects of network instability, environmental variability and inference randomness, we report the average performance over three runs.
To evaluate our automatic evaluator, we use 1,444 OSWorld trajectories generated by Claude-Sonnet-4, Claude-Sonnet-4.5, O3, and OpenCUA-7B, and report precision, recall, and agreement with the rule-based evaluator. To assess reliability during RL training, we further sample 288 trajectories across six environments and compute the same metrics against human judgments.

\begin{figure*}[!ht]
  \centering
    \includegraphics[width=0.95\textwidth]{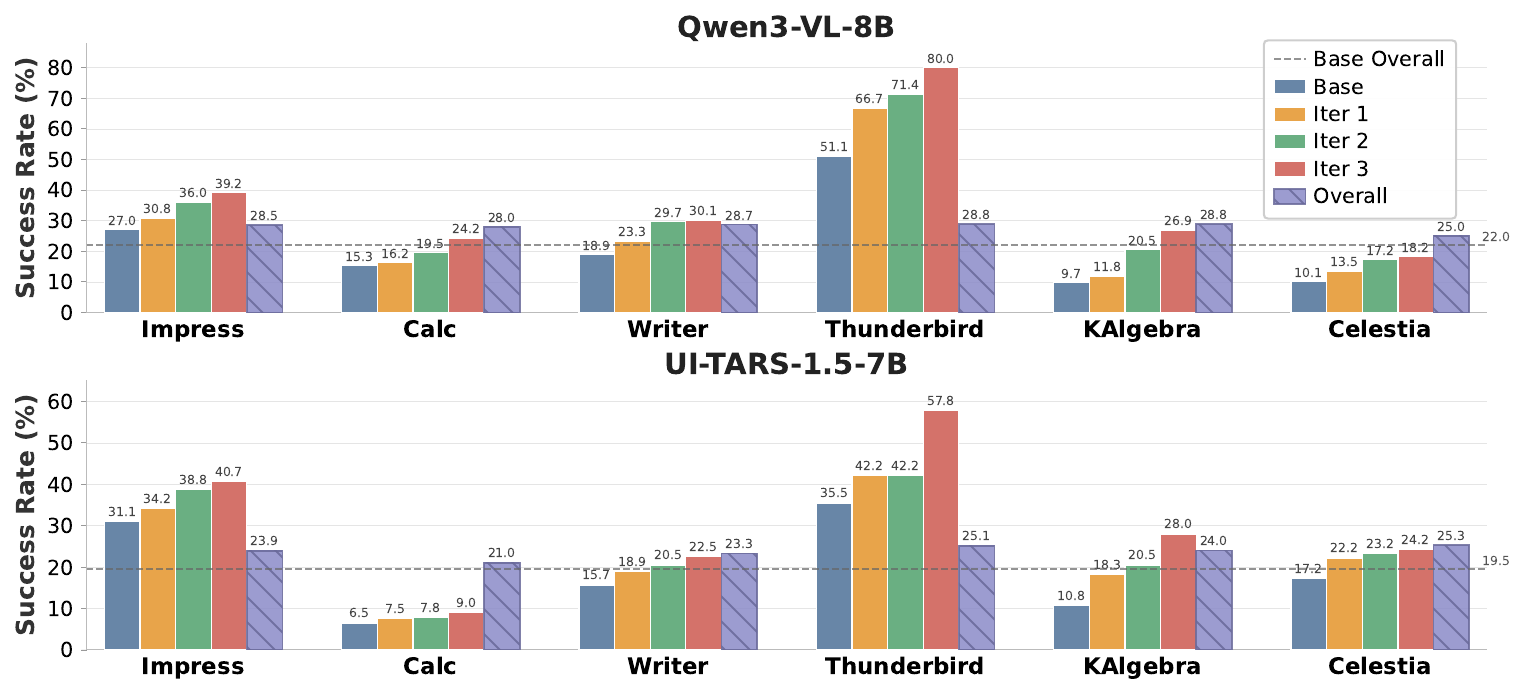}
    \caption{
        Average success rate across six environments for the base agent and agents that continually learn in specific environments via \ACuRL{}. The x-axis denotes the training environment, and the y-axis reports the corresponding evaluation performance. The overall score is the average performance across all six environments, with the gray dashed line indicating the base agent’s overall performance. “Iter” denotes the number of training iterations.
    }
    \label{fig:main_result_figure}
\end{figure*}
 
\noindent \textbf{Implementation Details.}
We adopt UI-TARS-1.5-7B~\citep{ui-tars-15-seed} and Qwen3-VL-8B-Instruct~\citep{bai2025qwen3vltechnicalreport} as base agents.
As UI-TARS-1.5-7B predates ScienceBoard, environments like Celestia and KAlgebra are likely unseen, thereby enabling us to evaluate the effectiveness of \ACuRL{} in novel environments.
For our task generator, we use GPT-5~\citep{openaigpt5}.
For each environment, we generate 256 tasks per iteration from web-crawled contexts. 
For CUAJudge, we use GPT-5-mini to reduce evaluation cost. 
See Appendix~\ref{appendix:experimental_details} for more implementation details.
\subsection{Infrastructure}
Building stable, efficient infrastructure for CUAs poses unique challenges. Each rollout requires an isolated Linux environment to prevent cross-trajectory interference, leading to high resource overhead.
Moreover, environment initialization and response latency directly affect training throughput. To support scalable, high-throughput training, we introduce a series of infrastructure-level optimizations that accelerate training speed by 3–5 times based on our preliminary exploration. 
\begin{itemize}[leftmargin=*, topsep=1pt, itemsep=1pt, parsep=0pt]
    \vspace{-1mm}
    \item \textbf{Unified environment management protocol.}
    We introduce a lightweight API protocol that enables batch environment creation and deletion via IP-based requests, with built-in failure detection and automatic restarts. This design greatly simplifies environment orchestration and facilitates efficient large-scale environment scaling.
    \item \textbf{Environment preloading and asynchronous evaluation.}
    Environment initialization and trajectory evaluation are two major bottlenecks to training speed. Each step requires reinitializing environments, which typically takes 5--10 minutes. As a result, GPUs remain idle and stall training. To address this issue, we introduce environment preloading: while the agent is being updated, a separate worker asynchronously initializes environments for the next step and deletes previous environments. 
    Furthermore, instead of synchronously evaluating trajectories after all rollouts complete, we deploy a separate evaluation worker that continuously monitors and asynchronously evaluates trajectories upon completion, thereby further improving overall training efficiency.
\end{itemize}

\subsection{Intra-Environment Continual Learning}
As shown in Figure~\ref{fig:main_result_figure}, the agent achieves progressive gains of 3–29\% on target environments across iterations, demonstrating effective intra-environment continual learning with autonomous skill acquisition on increasingly complex tasks.
Importantly, performance on other environments is also preserved or even improved, yielding overall gains of 2–7\% without catastrophic forgetting. Detailed results on each environment after continual learning are provided in Appendix~\ref{appendix:detailed_results_for_main_figure}.
We attribute these gains to positive cross-environment knowledge transfer, where agents acquire reusable interaction skills and decision patterns that generalize across environments (e.g., Impress to Writer).
We provide a detailed analysis of these phenomena in Section~\ref{sparsity} and Appendix~\ref{appendix:parameter_update_overlap}.

\subsection{Cross-Environment Continual Learning}
\input{tables/continual_learning_multiple_envs_qwen3vl}
To evaluate the generalization of \ACuRL{} to the cross-environment continual learning scenario, we conduct experiments where agents sequentially learn 2-5 distinct environments. Specifically, we initialize an agent already adapted to one environment and then perform an additional round of continual learning to enable adaptation to different target environments.
As shown in Table~\ref{tab:cross_env}, it remains effective as the number of environments increases, improving target performance and achieving the highest overall score. We conjecture that gains across seemingly unrelated environments arise from transferable decision patterns and shared interface-level knowledge, such as general exploration strategies, avoidance of redundant actions, and common UI conventions.
Overall, these results demonstrate that our curriculum-based continual learning strategy facilitates effective knowledge accumulation while mitigating catastrophic forgetting. Moreover, it generalizes beyond single-environment specialization and scales to multiple environments, highlighting its suitability for realistic continual learning scenarios. Although \ACuRL{} achieves substantial improvements, the agent’s performance remains far below human level. To better understand this gap, we conduct a detailed failure case analysis. We find that the majority of errors stem from knowledge and planning limitations, premature termination, and repetitive action loops. See Appendix~\ref{appendix:failure_case_analysis} for more details.

\subsection{Effectiveness of ACuRL in Mitigating Environment Changes}
As illustrated in Figure~\ref{fig:dynamic_figure}, real-world environment changes such as platform migration, software updates, UI modifications, and resolution shifts lead to substantial performance drops of up to 51\% even on the same test set. This observation suggests that current CUAs are highly sensitive to environmental dynamics and underscores the necessity of continual learning for adaptation. To demonstrate that \ACuRL{} can effectively mitigate these effects, we perform three iterations of continual training with \ACuRL{}. 
Results show that \ACuRL{} substantially improves performance in post-change environments. For example, it achieves a 145\% improvement under platform migration from Ubuntu to Windows, as well as gains of up to 145\% under version updates and 77\% under resolution shifts. Overall, these findings indicate that \ACuRL{} enables effective adaptation to diverse dynamic changes in real-world digital environments, supporting the practical deployment of CUAs.
\begin{figure}[!ht]
  \centering
    \includegraphics[width=\linewidth]{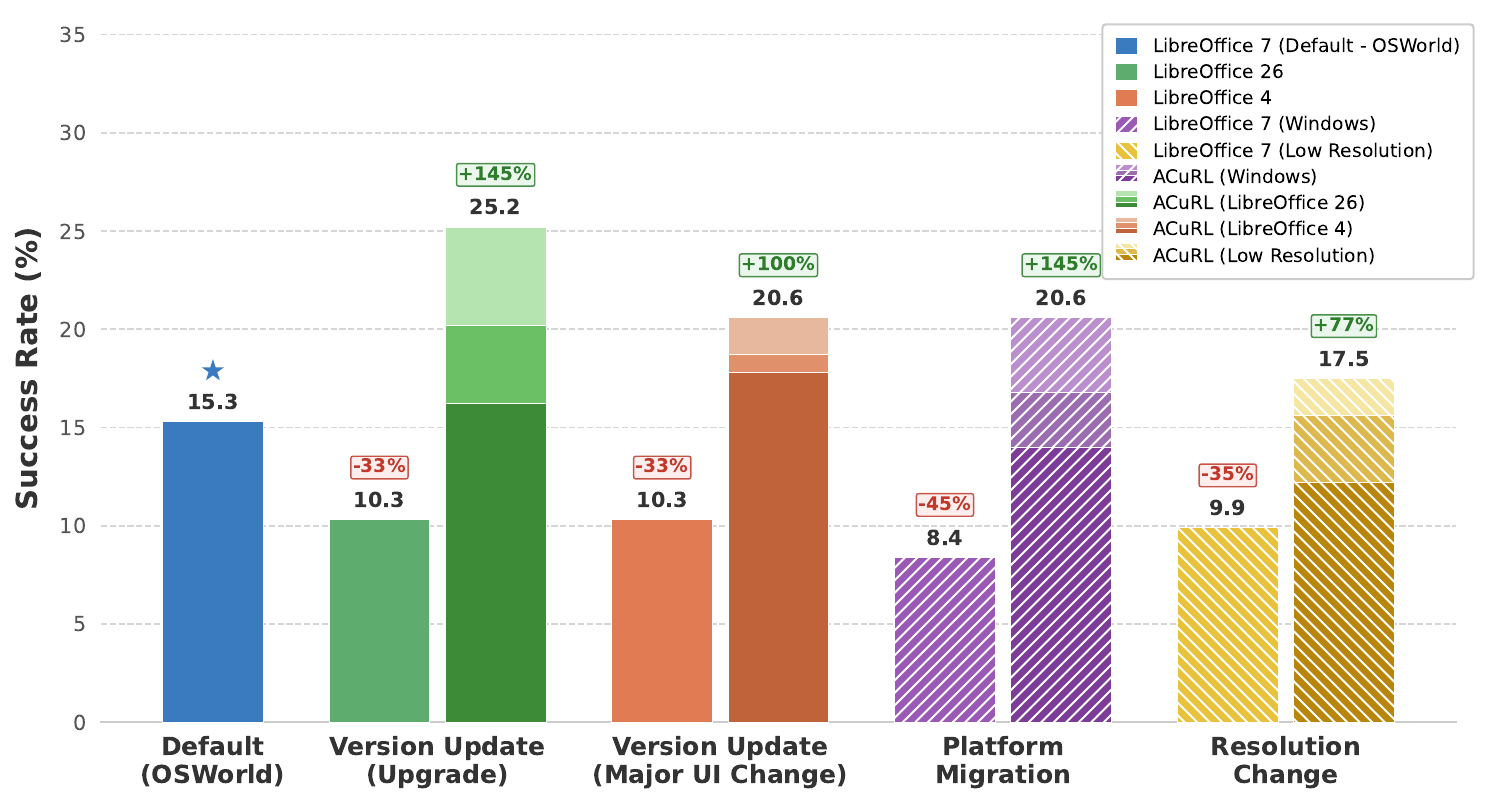}
    \caption{
      Effectiveness of ACuRL under environmental dynamics on Calc. It mitigates performance drops and improves performance across software updates, platform migration, and resolution changes. Color intensity indicates iterations (darker = earlier, lighter = later).
    }
    \label{fig:dynamic_acurl}
    \vspace{-6mm}
\end{figure}

\input{tables/judge_comparison_table}

\subsection{Ablation Study}
To assess the contribution of each component, we conduct ablation studies that remove iterative training and curriculum learning during training.
For iterative RL training, we train the agent for the same number of optimization steps using only tasks from the first iteration (75, 150 and 225 steps corresponding to Iter 1, 2 and 3).
To ablate curriculum learning, we replace our task generator with random task generation at each iteration, without using evaluation feedback from the previous iteration.
As shown in Table~\ref{tab:ablation}, removing either iterative RL training or curriculum-based task generation leads to substantial performance degradation. Specifically, without iterative training, performance peaks at 75 training steps (i.e., Iter 1) and then declines. This is because training on a fixed task pool induces overfitting, thereby impairing generalization to unseen tasks.
In addition, replacing curriculum-based task generation with random generation yields smaller gains.
These results highlight the importance of curriculum-driven task generation, where task difficulty is dynamically aligned with the agent’s evolving capabilities to enable effective continual learning. We further compare \ACuRL{} with recent curriculum-based methods and find that it achieves competitive or superior performance across environments; see Appendix~\ref{appendix:comparison_baselines} for details. While GPT-5 is used as the generator for its strong instruction-following ability, the gains are not attributable to a stronger external model, as GPT-5 lacks GUI navigation capabilities. Replacing it with the open-source Qwen3-VL-8B still yields substantial improvements. Detailed results are provided in Appendix~\ref{appendix:impact_generator}.
\input{tables/ablation}
\subsection{Further Analysis}
\noindent \textbf{The Reliability of CUAJudge.}
The effectiveness of continual learning through RL critically depends on the reliability of the automatic evaluator. To this end, we compare \CUAJudge{} with the rule-based evaluator on 1,444 publicly released trajectories. As shown in Table \ref{tab:judge_comparison}, \CUAJudge{} achieves a high agreement of 87.5\% with the rule-based evaluator. Furthermore, compared with the state-of-the-art evaluator~\citep{xue2025an,lu2025agentrewardbench}, \CUAJudge{} improves precision by 4.1\% across trajectories from four different agents, substantially reducing false positives. As the state difference analysis module is a key component of \CUAJudge{}, we conduct an ablation study to evaluate its importance; see Appendix~\ref{appendix:details_cuajudge} for details.
To further verify that \CUAJudge{} can provide reliable reward signals for training tasks and on-policy trajectories, we conduct a human evaluation on 288 trajectories. Specifically, we collect 48 trajectories per environment, evenly sampled across three iterations from six environments. As shown in Table~\ref{tab:judge_with_human}, \CUAJudge{} consistently exhibits high agreement with human judgments (93.7\%). Notably, as training progresses across iterations, precision and agreement remain consistently around 90\%, suggesting that \CUAJudge{} avoids exploitable shortcuts for reward hacking and remains well aligned with true task completion.
Overall, these findings suggest that \CUAJudge{} could serve as a robust and reliable automatic evaluator of CUA trajectories for many learning paradigms, including Rejection Sampling Fine-Tuning (RFT)~\citep{yuan2023scalingrelationshiplearningmathematical}, Reflection~\citep{shinn2023reflexion,paul-etal-2024-refiner,xue2023rcotdetectingrectifyingfactual}, and RL.
\input{tables/human_eval_judge}

\noindent \textbf{The Sparsity of Agent Updates during Continual Learning.}
\label{sparsity} As shown in Table~\ref{tab:specialist_models} and Table~\ref{tab:cross_env}, the agent achieves substantial gains on target environments under both intra- and cross-environment continual learning, while maintaining stable or even improved performance on other environments without catastrophic forgetting.
Although~\citep{mukherjee2025reinforcement, chen2025retainingdoingroleonpolicy} observe sparse parameter updates during RL, their studies are limited to language models, leaving it unclear how parameters update for multimodal CUAs, where decision-making is heavily driven by visual perception from dense, information-rich screenshots.
Therefore, to better understand how the model’s internal representations evolve during continual adaptation, we analyze model sparsity across different iterations.
Formally, we quantify it using \emph{update sparsity}~\citep{mukherjee2025reinforcement}, which measures the fraction of parameters that remain effectively unchanged.
Let $\theta^0, \theta^1 \in \mathbb{R}^n$ denote the model parameters before and after fine-tuning, respectively, where $n$ is the total number of trainable parameters.
Given a tolerance $\varepsilon$ to ignore negligible numerical variations, we set $\varepsilon = 1\times10^{-5}$ following \citet{mukherjee2025reinforcement}, and define \emph{update sparsity} as
\begin{equation}
\mathrm{UpdateSparsity}_{\varepsilon}(\theta^0,\theta^1)
= 1 - \frac{1}{n}\sum_{i=1}^{n}
\mathbb{I}\!\left(|\theta_i^1-\theta_i^0|>\varepsilon\right),
\end{equation}

\begin{figure}[h]
    \centering
    \begin{minipage}{0.48\linewidth}
        \centering
        \includegraphics[width=\linewidth]{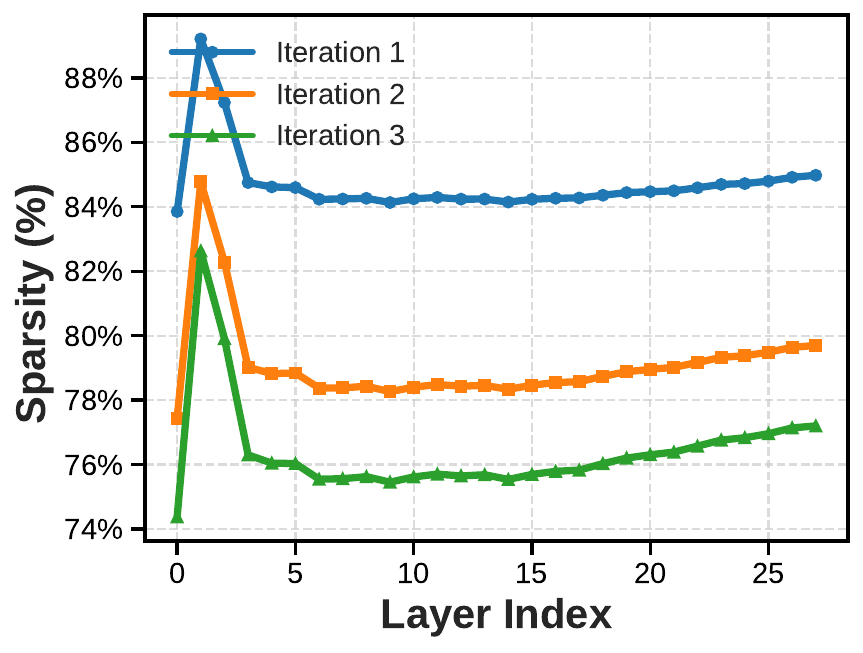}
        \subcaption{LLM backbone}
        \label{fig:sparsity-llm}
    \end{minipage}
    \hfill
    \begin{minipage}{0.48\linewidth}
        \centering
        \includegraphics[width=\linewidth]{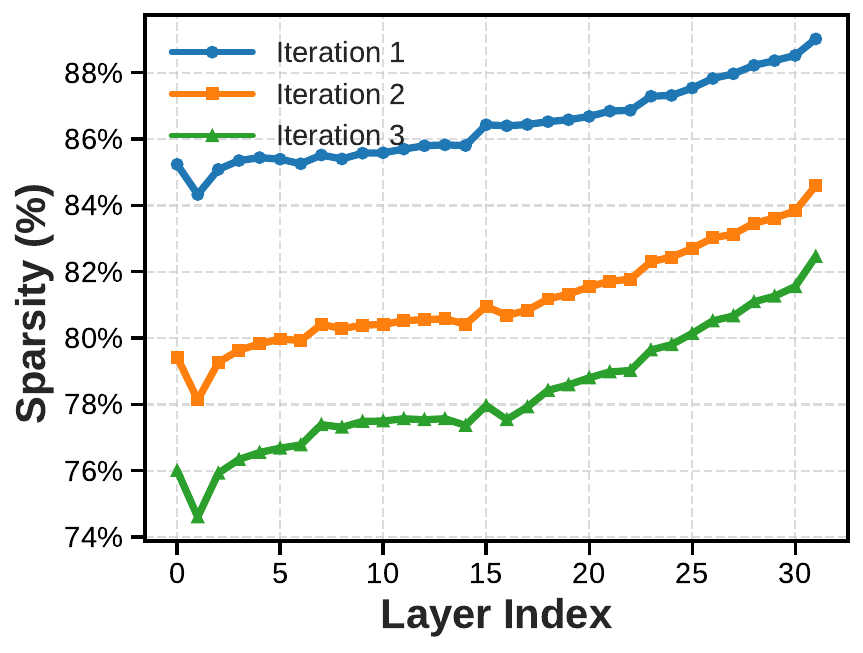}
        \subcaption{Vision encoder}
        \label{fig:sparsity-vision}
    \end{minipage}
    \caption{Sparsity of parameter updates across RL iterations, averaged over six environments.}
    \label{fig:sparsity}
    \vspace{-5mm}
\end{figure}

As shown in Figure~\ref{fig:sparsity}, although sparsity gradually decreases over training, both the LLM backbone and vision encoder remain highly sparse, with around 80\% of parameters unchanged. This helps explain why the agent can continually learn in target environments while preserving performance on others.
Moreover, the two components exhibit distinct layer-wise patterns: the LLM backbone maintains relatively uniform sparsity across middle and upper layers (Layers~3–27), whereas the vision encoder becomes increasingly sparse with depth.
We also observe a striking contrast in the early layers of the two modules. The first one to two layers of the LLM backbone exhibit unusually high sparsity, whereas the first layer of the vision encoder shows much lower sparsity. 
This asymmetry suggests that early vision layers, responsible for low-level visual perception, require stronger adaptation in new environments, while early LLM layers are less involved in environment-specific perception or decision making. We leave understanding why RL yields sparse updates and how this naturally mitigates forgetting to future work.

%% file: tables/continual_learning_multiple_envs_qwen3vl.tex
\begin{table*}[h!]
\centering
\small
\caption{The performance of continual learning in the cross-environment scenario where the agent sequentially learns more than one environment. Rows correspond to agents continually learning in one or multiple environments and columns to testing environments. Results are grouped by backbone models. The best results are \textbf{bolded}. The overall score is the mean performance over all six environments.}
\label{tab:cross_env}
\resizebox{\textwidth}{!}{
\begin{tabular}{llccccccc}
\toprule
\textbf{Model} & \textbf{Agent} & \textbf{Impress} & \textbf{Calc} & \textbf{Writer} & \textbf{Thunderbird} & \textbf{KAlgebra} & \textbf{Celestia} & \textbf{Overall} \\
\midrule

\multirow{7}{*}{\textbf{UI-TARS-1.5-7B}}
& Base Agent
& 31.1 & 6.5 & 15.7 & 35.5 & 10.8 & 17.2 & 19.5 \\

& Impress
& \textbf{40.7} & 5.9 & 20.9 & 40.0 & 12.9 & 23.2 & 23.9 \\

& Calc
& 33.2 & \textbf{9.0} & 16.9 & 37.8 & 16.1 & 13.1 & 21.0 \\

& KAlgebra
& 30.8 & 5.3 & 16.9 & 48.9 & \textbf{28.0} & 14.2 & 24.0 \\

\rowcolor{gray!15}
& Impress $\rightarrow$ Calc
& 37.0 & 8.7 & \textbf{23.7} & 44.5 & 16.1 & 19.2 & 24.9 \\

\rowcolor{gray!15}
& Impress $\rightarrow$ KAlgebra
& 36.7 & 6.8 & 21.7 & 46.7 & 18.3 & 20.2 & 25.1 \\

\rowcolor{gray!15}
& Impress $\rightarrow$ KAlgebra $\rightarrow$ Calc
& 33.9 & 6.6 & 23.3 & \textbf{51.1} & 16.1 & \textbf{24.2} & \textbf{25.9} \\

\midrule

\multirow{11}{*}{\textbf{Qwen3-VL-8B}}
& Base Agent
& 27.0 & 15.3 & 18.9 & 51.1 & 9.7 & 10.1 & 22.0 \\

& Impress
& \textbf{39.2} & 14.0 & 24.9 & 64.5 & 17.2 & 11.1 & 28.5 \\

& Calc
& 33.0 & \textbf{24.2} & 26.5 & 62.2 & 10.8 & 11.1 & 28.0 \\

& Writer
& 27.7 & 17.1 & \textbf{30.1} & 68.9 & 12.9 & 15.2 & 28.7 \\

& KAlgebra
& 30.1 & 17.1 & 28.5 & 62.2 & \textbf{26.9} & 8.1 & 28.8 \\

& Celestia
& 26.4 & 12.2 & 18.1 & 62.2 & 12.9 & \textbf{18.2} & 25.0 \\

\rowcolor{gray!15}
& Impress $\rightarrow$ Calc
& 35.9 & 15.5 & 22.9 & 70.0 & 17.8 & 13.7 & 29.3 \\

\rowcolor{gray!15}
& Impress $\rightarrow$ KAlgebra
& 35.8 & 15.9 & 19.7 & \textbf{71.1} & 20.5 & 11.1 & 29.0 \\

\rowcolor{gray!15}
& Impress $\rightarrow$ KAlgebra $\rightarrow$ Calc
& 35.4 & 19.7 & 18.1 & 68.9 & 22.6 & 14.2 & 29.8 \\

\rowcolor{gray!15}
& Impress $\rightarrow$ KAlgebra $\rightarrow$ Calc $\rightarrow$ Celestia
& 33.9 & 19.3 & 22.5 & \textbf{71.1} & 25.8 & 16.2 & 31.5 \\

\rowcolor{gray!15}
& Impress $\rightarrow$ KAlgebra $\rightarrow$ Calc $\rightarrow$ Celestia $\rightarrow$ Writer
& 37.3 & 18.7 & 25.7 & 66.7 & 24.7 & 17.2 & \textbf{31.7} \\

\bottomrule
\end{tabular}
}
\end{table*}

%% file: tables/judge_comparison_table.tex
\begin{table*}[ht!]
\centering
\small
\caption{Comparison of WebJudge and CUAJudge across trajectories generated by different agents.}
\resizebox{\textwidth}{!}{
\begin{tabular}{l|ccc|ccc|ccc|ccc|c}
\toprule
 & \multicolumn{3}{c|}{\textbf{Claude-Sonnet-4}}
 & \multicolumn{3}{c|}{\textbf{Claude-Sonnet-4.5}}
 & \multicolumn{3}{c|}{\textbf{O3}}
 & \multicolumn{3}{c|}{\textbf{OpenCUA-7B}}
 & \textbf{Overall} \\
\midrule
\textbf{Evaluator}
 & Precision & Recall & Agreement
 & Precision & Recall & Agreement
 & Precision & Recall & Agreement
 & Precision & Recall & Agreement
 & Agreement \\
\midrule
WebJudge
 & 71.7 & 82.7 & 84.8
 & 79.4 & 84.1 & 84.2
 & 38.6 & \textbf{53.1} & 88.4
 & 59.0 & 71.0 & 85.0
 & 85.6 \\
CUAJudge
 & \textbf{74.8} & \textbf{86.4} & \textbf{87.0}
 & 82.1 & \textbf{84.8} & \textbf{85.9}
 & 42.1 & 50.0 & 89.4
 & 66.2 & 71.0 & 87.5
 & \textbf{87.5} \\
\bottomrule
\end{tabular}
}
\label{tab:judge_comparison}
\end{table*}

%% file: tables/ablation.tex
\begin{table}[!h]
\centering
\small
\caption{Ablation study on the impact of iterative training and curriculum learning
in the LibreOffice Impress environment.}
\label{tab:ablation}
\resizebox{\linewidth}{!}{
\begin{tabular}{lccc}
\toprule
\textbf{Method} & \textbf{Iter 1} & \textbf{Iter 2} & \textbf{Iter 3} \\
\midrule
Iterative Curriculum RL 
& \textbf{34.2} & \textbf{38.8} & \textbf{40.7} \\
\hspace{2mm} - w/o Iterative Training 
& 34.2 \textcolor{red}{(-0.0)} & 34.0 \textcolor{red}{(-4.8)} & 32.6 \textcolor{red}{(-8.1)} \\
\hspace{2mm} - w/o Curriculum Learning 
& 29.8 \textcolor{red}{(-4.4)} & 33.6 \textcolor{red}{(-5.2)} & 36.4 \textcolor{red}{(-4.3)} \\
\bottomrule
\end{tabular}
}
\end{table}

%% file: tables/human_eval_judge.tex
\begin{table}[t]
\centering
\small
\caption{\CUAJudge{}'s precision, recall, and agreement with human judgments.}
\label{tab:judge_with_human}
\begin{tabular}{lccc}
\toprule
\textbf{Iteration} & \textbf{Precision} & \textbf{Recall} & \textbf{Agreement} \\
\midrule
Iter 1    & 98.0 & 96.2 & 96.6 \\
Iter 2    & 89.1 & 89.1 & 89.1 \\
Iter 3    & 95.8 & 95.8 & 95.6 \\
\rowcolor{gray!15}
Overall  & 94.5 & 93.8 & 93.7 \\
\bottomrule
\end{tabular}
\end{table}

%% file: sections/conclusion.tex
\section{Conclusion}
In this work, we propose a fully autonomous curriculum reinforcement learning framework (\ACuRL{}) that enables CUAs to continually adapt to specific environments in a zero-data setting.
By grounding task generation in accumulated environment experience and progressively adjusting task difficulty to match the agent’s evolving capabilities, \ACuRL{} consistently improves performance in both intra- and cross-environment continual learning across six representative environments, without catastrophic forgetting. Moreover, it effectively mitigates performance degradation under diverse environmental dynamics, including version updates, platform migration, and resolution changes.
Further analysis reveals that such adaptation is driven by highly sparse parameter updates, offering insights into how agents acquire new skills while preserving prior knowledge. Overall, these results establish \ACuRL{} as a scalable and effective paradigm for continual learning in complex computer-use environments.

%% file: sections/acknowledgments.tex
\section*{Acknowledgments}
The authors would thank colleagues from the OSU NLP group for constructive feedback. This research was sponsored in part by NSF CAREER \#1942980, NSF CAREER \#2443149, the Alfred P. Sloan Research Fellowship, and gifts from Cisco and Amazon. We also appreciate computational resources provided by the Ohio Supercomputer Center~\citep{Ohio_Supercomputer_Center1987-dl}. The views and conclusions contained herein are those of the authors and should not be interpreted as representing the official policies, either expressed or implied, of the U.S. government. The U.S. Government is authorized to reproduce and distribute reprints for Government purposes notwithstanding any copyright notice herein.

%% file: sections/appendix.tex
\clearpage
\appendix

\crefname{section}{Appendix}{Appendices}
\newpage
\onecolumn
\textbf{Table of Content:}
\begin{enumerate}[nosep]
    \item {\cref{appendix:related_work}: Related Work}
    \item {\cref{appendix:limitations}: Limitations}
    \item {\cref{appendix:impact_statements}: Impact Statements}
    \item {\cref{appendix:background}: Background}
    \item {\cref{appendix:details_curriculum_task_generation}: Details of Curriculum Task Generation}
    \item {\cref{appendix:details_cuajudge}: Details of CUAJudge}
    \item {\cref{appendix:task_validity}: Task Validity Analysis}
    \item {\cref{appendix:parameter_update_overlap}: The Parameter Update Overlap across Agents}
    \item {\cref{appendix:experimental_details}: Experimental Details}
    \item {\cref{appendix:comparison_baselines}: Comparison with Existing Methods}
    \item {\cref{appendix:failure_case_analysis}: Failure Case Analysis}
    \item {\cref{appendix:detailed_results_for_main_figure}: Detailed Results for Each Environment}
    \item {\cref{appendix:prompts}: Prompts}
    \item {\cref{appendix:example_contexts}: Examples of Context Across Different Environments}
\end{enumerate}
\newpage

\twocolumn
\appendix

\clearpage

\input{sections/related_work}

\input{sections/limitations}

\input{sections/impact_statements}

\input{sections/background}

\section{Details of Curriculum Task Generation}
\label{appendix:details_curriculum_task_generation}

\subsection{Iteration 0}
Notably, under our zero-data setting, no tasks are available at the beginning of training to assess the agent’s initial capability. To address this, we let the generator $G$ synthesize an initial task set based on context and exploration trajectories. We then conduct a capability evaluation on this task set to obtain feedback, which is subsequently used to generate the task set for Iteration 1, explicitly tailored to the current agent. We refer to this initialization phase as Iteration 0.
During Iteration 0, the task generator focuses on synthesizing a set of atomic tasks that cover essential and frequently used atomic actions in target environments, thereby providing a foundational skill set for subsequent curriculum learning. Tasks from iteration 0 are not used for training; they are only used to assess the agent’s initial capabilities and guide task synthesis in later iterations.

\subsection{Medium Level Adjustment}
Tasks at a medium level suggest that the agent has not yet fully mastered the underlying skills required and can therefore continue to learn from them, expanding its edge of competence~\citep{zhang2025interplaypretrainingmidtrainingrl}. Accordingly, to maintain task diversity while supporting continual learning and skill consolidation, we let the curriculum generator produce tasks that exercise the same or similar skills across varied contexts, user personas, and task goals, but are realistic (e.g., completing the same document-editing operation under different constraints, such as layout requirements or formatting standards, or plotting charts based on different data.).

\subsection{Context Reviews}
\label{app:context_review}
For each environment, we collect diverse and substantially different contextual materials from the web. For example, in the LibreOffice Impress and Writer environments, we gather 1–3 contexts from multiple domains, including art, technology, social sciences, education, and scientific research. For environments where contexts are difficult to obtain or do not exist (e.g., KAlgebra), we instead let the agent explore the environment directly to acquire environment experience. From each context, we generate 32 tasks. The prompt for context review is ``\textit{You are given access to a software application in its current state. Your task is to actively and read-only explore the application to understand the full user context, including multiple distinct content areas, views, or sections, not just a single default view. You must navigate into and inspect representative content from each major content-bearing area (e.g., different folders, tabs, panels, lists, or sections) to verify what information exists and how it differs across areas. Do not modify, create, delete, send, save, or trigger any actions that persist changes. You must not finish after exploring only one section or view; only finish after you have reasonably covered the breadth of the application’s content space.}''.

\subsection{Task Evolving Across Different Iterations}
\begin{figure}[ht]
  \centering
  \begin{subfigure}[t]{\columnwidth}
    \centering
    \includegraphics[width=\linewidth]{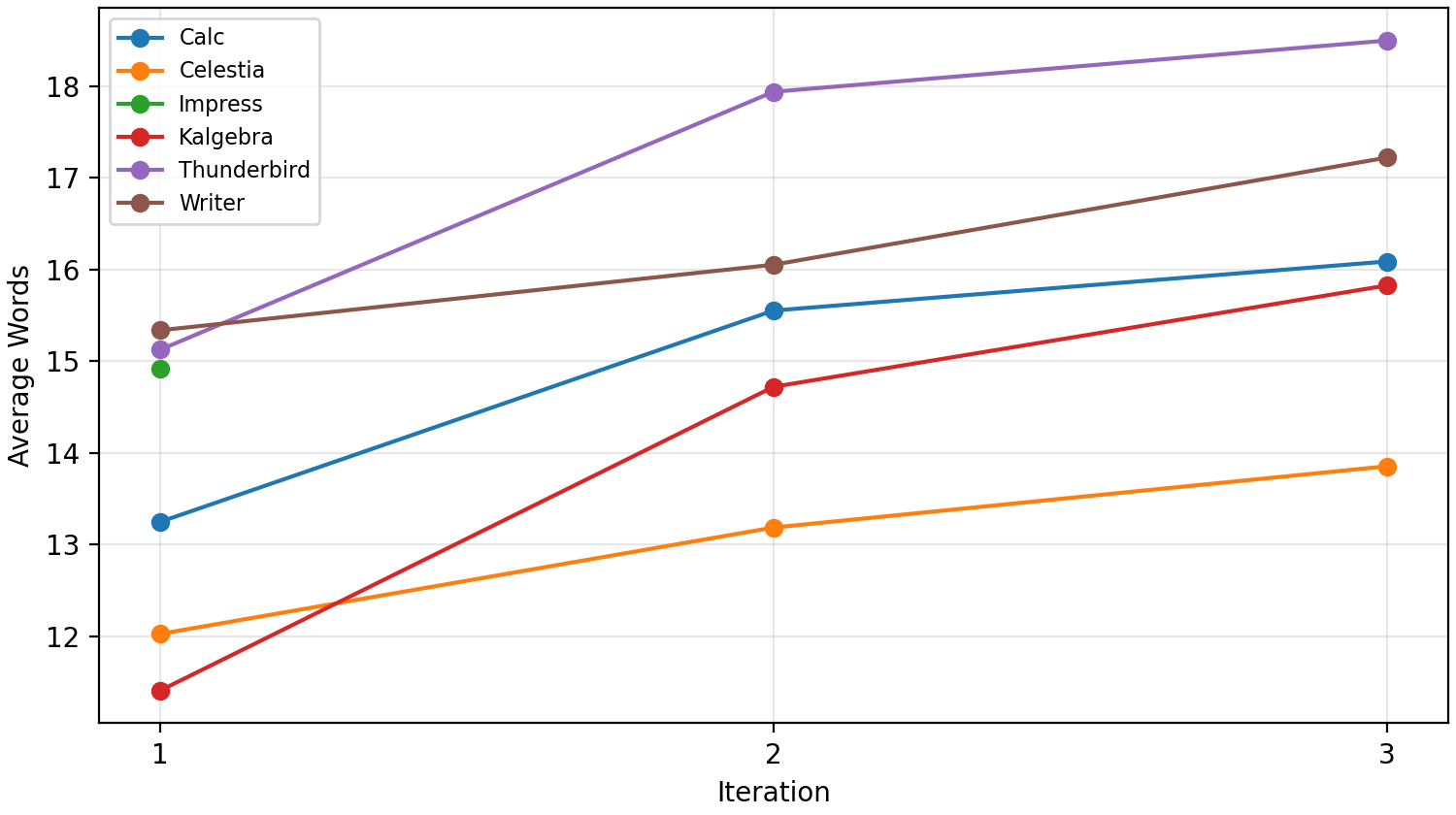}
    \caption{Average instruction length across iterations.}
    \label{fig:task_length}
  \end{subfigure}
  \hfill
  \begin{subfigure}[t]{\columnwidth}
    \centering
    \includegraphics[width=\linewidth]{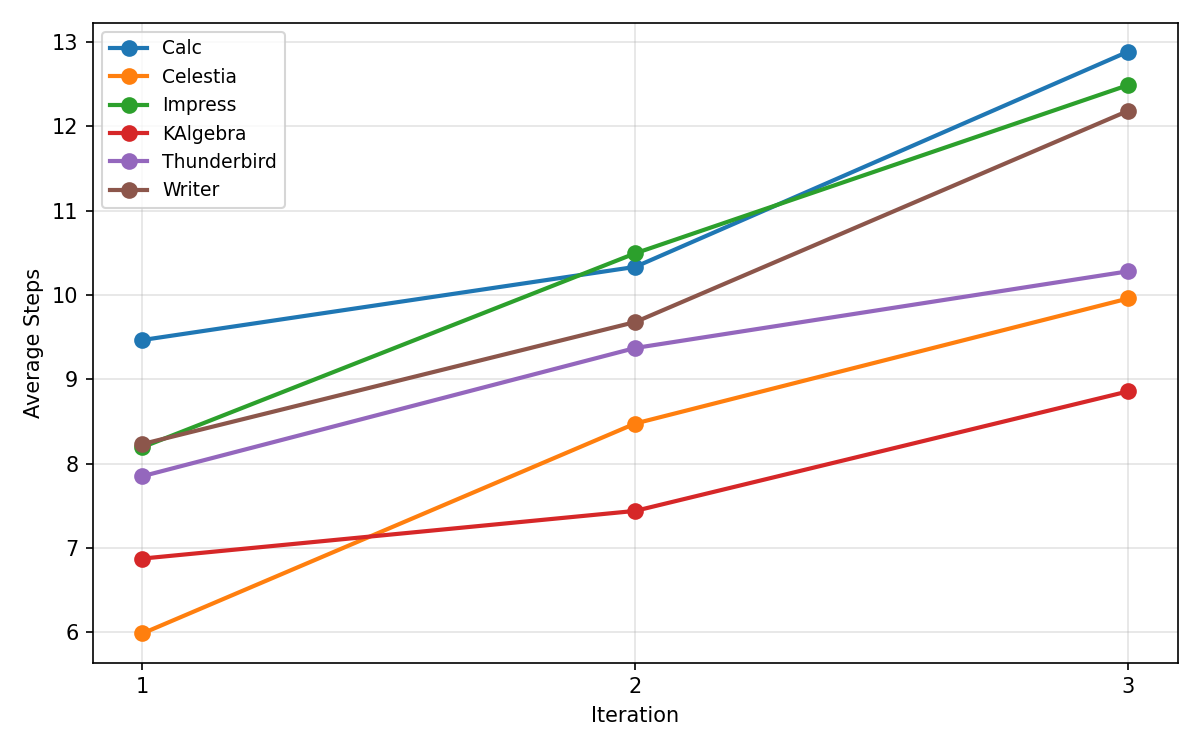}
    \caption{Average execution steps across iterations.}
    \label{fig:task_execution_step}
  \end{subfigure}
  \caption{
  Task complexity increases across iterations in different environments.
  }
  \label{fig:task_complexity}
\end{figure}

To better illustrate the quality of the synthesized tasks and how they evolve across iterations, we analyze both the average task length and the number of steps taken by the agent to complete tasks throughout training. As shown in Figure ~\ref{fig:task_length} and ~\ref{fig:task_execution_step}, both metrics exhibit a consistent increasing trend over iterations, indicating that the generated tasks become progressively longer and, to some extent, more complex. This behavior aligns with our expectations: as iterative training improves the agent’s capabilities, the curriculum generator produces increasingly challenging tasks that are better tailored to the agent’s current ability. We further conduct a qualitative analysis and present representative examples in Table ~\ref{tab:task_example}. These examples demonstrate that task complexity increases across iterations, evolving from operations on a single object to interactions involving multiple objects, and from requiring a single skill to composing multiple skills within a single task.

\subsection{The impact of curriculum generator}
\label{appendix:impact_generator}
\input{tables/geneator_comparison}
Some agents are specialized for GUI navigation and thus lack strong instruction-following capabilities, which prevents them from generating valid tasks or supporting curriculum-based learning (e.g., UI-TARS). In \ACuRL{}, we instead adopt GPT-5 as the generator due to its strong instruction-following ability. However, this does not imply that our performance gains rely on a more powerful external agent, as GPT-5 itself does not possess GUI navigation capabilities. To further assess the impact of the generator, we replace GPT-5 with Qwen3-VL-8B, an open-source model with strong instruction-following ability. As shown in Table~\ref{tab:generator_impress}, \ACuRL{} remains highly effective, demonstrating that the observed gains do not stem from the frontier capabilities of GPT-5.

\section{Details of CUAJudge}
\label{appendix:details_cuajudge}

Since the number of API calls in the key screenshot identification stage of \CUAJudge{} scales linearly with the number of screenshots in a trajectory, the associated cost increases substantially. To reduce the overall cost during reinforcement learning, we therefore compare the performance and cost trade-offs of using GPT-5-mini and Qwen3-VL-8B. As shown in Table~\ref{tab:model_cost_precision}, using Qwen3-VL-8B for the key screenshot identification stage achieves higher precision while reducing the cost by approximately 5×, incurring only \$0.40 for evaluating 107 trajectories during RL training.

\input{tables/CUAJudge_key_screenshot_identification_trade_off}
We also conduct an ablation study to assess the role of the state difference analysis module. Specifically, we remove the component that compares the initial and final states in \CUAJudge{} and exclude it from the outcome judgment. As shown in Table~\ref{tab:judge_comparison_ablation}, removing this module leads to clear declines in both recall and overall agreement, highlighting its importance.
\input{tables/judge_ablation}

\section{Task Validity Analysis}
\label{appendix:task_validity}
To quantify the quality of the synthesized tasks, we randomly sample 144 tasks across three training iterations from six different environments. We find that only 8 tasks are invalid, while 94\% of the tasks are valid and explicitly grounded in environment-specific contexts. These results demonstrate that our task synthesis pipeline can scalably generate high-quality, environment-grounded tasks for target environments without relying on any human supervision. The eight invalid tasks primarily arise from ambiguous task descriptions, incomplete information, or cases where the environment does not provide the required functionality to support the specified task.
\section{The Parameter Update Overlap across Agents}
\label{appendix:parameter_update_overlap}
\begin{figure}[h!]
  \vskip 0.2in
  \centering
  \begin{subfigure}[t]{0.48\columnwidth}
    \centering
    \includegraphics[width=\linewidth]{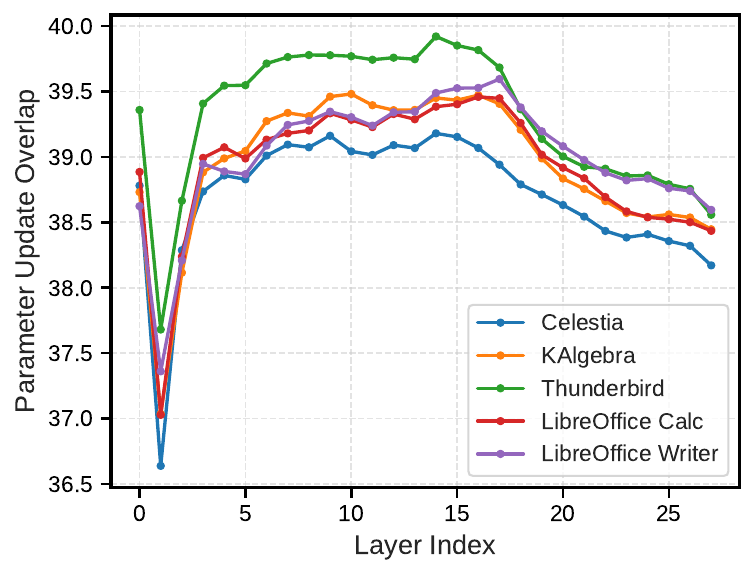}
    \caption{LLM backbone}
    \label{fig:overlap-llm}
  \end{subfigure}
  \hfill
  \begin{subfigure}[t]{0.48\columnwidth}
    \includegraphics[width=\linewidth]{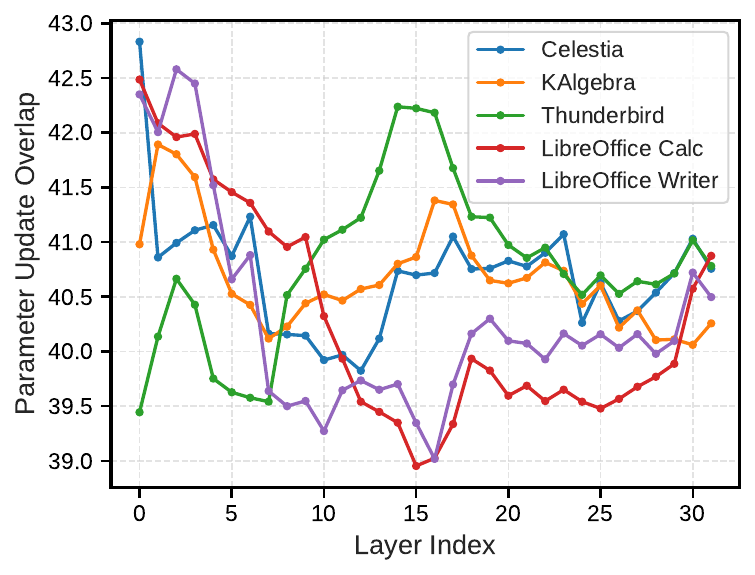}
    \caption{Vision encoder}
    \label{fig:overlap-vision}
  \end{subfigure}
  \caption{Overlap ratio of significantly updated parameters between LibreOffice Impress and other environments across different layers.}
  \label{fig:overlap}
\end{figure}

To better understand why continual learning in one target environment also improves performance in others, we conduct a fine-grained analysis of the similarity in internal parameter updates across agents undergoing continual learning in different environments. Specifically, we compute the pairwise overlap ratio of significantly updated parameters between the LibreOffice agent and agents trained on other environments. This overlap ratio quantifies the proportion of parameters that are significantly updated in both agents, serving as an indicator of shared internal adaptation patterns across environments.
As shown in Figure~\ref{fig:overlap}, both the LLM and vision encoder exhibit a 35–45\% parameter overlap across different specialist agents. This high degree of overlap provides a plausible explanation for why adapting to a specific environment can also benefit other environments: despite differences in surface-level interfaces, these environments share underlying decision patterns and reusable skills that enable effective transfer. We also observe distinct layer-wise overlap patterns between the LLM and the vision encoder. For the LLM, the overlap increases steadily across the early layers, peaks in the middle layers, and then gradually declines in the higher layers. This trend suggests that the learning of common skills and underlying decision-making patterns is primarily concentrated in the middle layers. In contrast, the vision encoder exhibits a more irregular and less structured relationship between overlap and layer depth.

\section{Experimental Details}
\label{appendix:experimental_details}
\paragraph{Training}
The number of API calls for key screenshot identification in \CUAJudge{} scales linearly with the trajectory length. Therefore, we adopt Qwen3-VL-8B~\citep{bai2025qwen3vltechnicalreport} for this component, which not only further reduces cost but also achieves higher precision. See Appendix~\ref{appendix:details_cuajudge} for details.
For iterative reinforcement learning, we employ strict on-policy training with three iterations on 8 × H100 GPUs and use a server with 96 CPU cores and 384 GB memory for hosting the environments. 
Each iteration is trained for 75 steps using a batch size of 16 and a group size of 8, and a learning rate of $1\times10^{-6}$. For the capability evaluation stage, we maintain consistency with training by sampling 8 rollouts with 1.0 temperature and measuring the average success rate as feedback. The final checkpoint of each iteration is used to initialize the next iteration. Following \citet{yu2025dapo} and \citet{wang2025nemotron}, we remove the KL divergence and adopt a higher clipping ratio of 0.28. To mitigate context length limitations, we retain only the most recent two images in the action history of each trajectory. Our RL framework is implemented based on VeRL~\citep{sheng2024hybridflow,feng2025groupingroup}.
For GRPO, we sample 8 trajectories per group with a temperature of 1.0. For the evaluator, following~\citet{xue2025an}, we set the threshold for filtering key screenshots to 3. For categorizing task difficulty, we set $\delta_{\text{low}}$ and $\delta_{\text{high}}$ to 30\% and 70\%, respectively: tasks with an average success rate below 30\% are considered difficult for the agent, those above 70\% indicate tasks that the agent has largely mastered, and the remaining tasks are classified as medium difficulty. 
Our threshold choice follows~\citet{chen2025self}. In GRPO, advantages are computed relatively within a group, and learning is maximized when the success rate is around 50\%. Therefore, we aim to keep task difficulty near this level during training. With a group size of 8, 3/8 and 5/8 are the closest practical approximations.
As task difficulty increases alongside the agent’s evolving capabilities, we iteratively increase both the maximum interaction steps and the maximum sequence length. Specifically, we use 15 steps with a maximum length of 13,000 tokens for iteration 1, 20 steps with 15,000 tokens for iteration 2, and 25 steps with 17,000 tokens for iteration 3. 
Regarding cost, we train the agent on 8×H100 GPUs. For environment hosting, we only require 96 CPU cores and 384 GB RAM to stably host 128 virtual machines. The generator cost is about \$10 per iteration (i.e., \$30 per environment for 3 iterations). For evaluation, we replace the intermediate steps in CUAJudge with Qwen3-VL-8B, reducing the cost by 5×. It costs only \$0.4 to evaluate 107 trajectories, corresponding to about \$0.48 per training step. With 225 steps in total (75 per iteration), the overall evaluation cost is approximately \$108 per environment.
\paragraph{Evaluation}
During evaluation, we uniformly set the maximum number of environment interaction steps to 25. For the ScienceBoard evaluations on KAlgebra and Celestia, we make a minor adjustment. Specifically, some tasks require the agent to produce answers in a strict format (e.g., prefixed with ANS:). We observe that in many cases the model provides a correct answer but is marked incorrect solely due to formatting mismatches. While the official evaluation protocol addresses this issue by adding extra instructions to explicitly enforce the required output format, we intentionally avoid introducing such additional instructions in order to strictly align with the OSWorld setting and evaluate the agent’s performance without auxiliary guidance, where the agent receives only the task description and no additional guidance. To address this issue, we replace this small subset of format-sensitive tasks with an LLM-as-a-judge evaluation powered by GPT-5-mini, which is given the model's prediction and the ground-truth answer and determines correctness. We manually inspect a portion of these cases and find that the judgments are consistently accurate. Table~\ref{tab:task_statistics} presents the task statistics for each environment.
For human evaluation, we provide evaluators with the full trajectory, including all agent actions and screenshots. The evaluator explicitly flags any uncertain cases; for such instances, a second evaluator is introduced, and the final label is determined through discussion.

\input{tables/task_statistics}

\section{Comparison with Existing Methods}
\label{appendix:comparison_baselines}
We compare \ACuRL{} with recent curriculum-based approaches, including SEAgent~\citep{sun2025seagent}, WebRL~\citep{qi2025webrl}, and DigitalRL~\citep{bai2024digirltraininginthewilddevicecontrol}. As shown in Table~\ref{tab:main_comparison}, \ACuRL{} achieves competitive or superior performance across environments, with particularly large gains on Impress (+12.6\%) and Celestia (+9.0\%). Notably, \ACuRL{} achieves these improvements with substantially lower computational cost. For instance, SEAgent requires supervised fine-tuning on synthesized trajectories followed by approximately 1,000 RL steps, whereas \ACuRL{} attains superior performance using only 225 RL steps.
\input{tables/comparison_baselines}

\section{Failure Case Analysis}
\label{appendix:failure_case_analysis}
\begin{figure}[!h]
  \begin{center}
    \centerline{\includegraphics[width=\columnwidth]{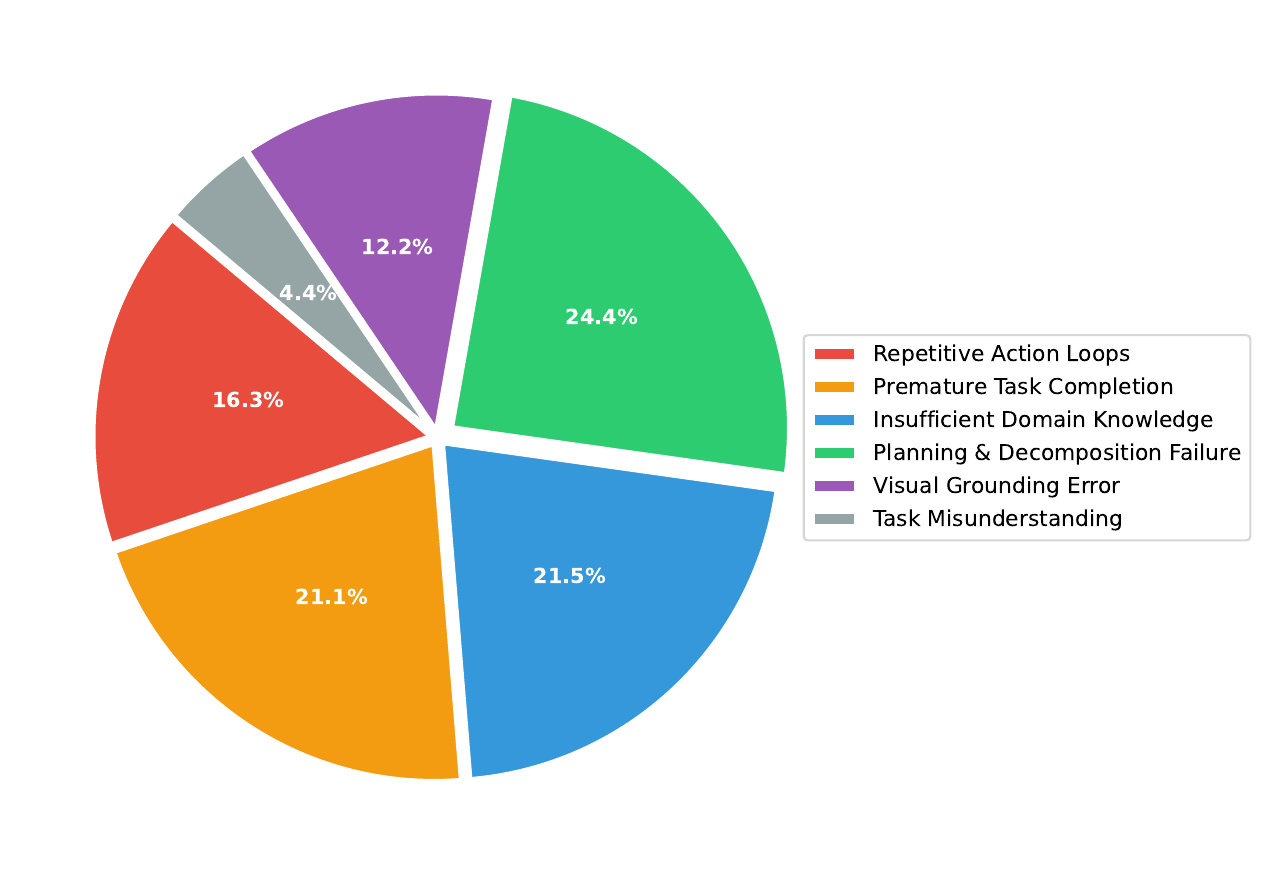}}
    \caption{
      Distribution of failure types based on root cause analysis.
    }
    \label{fig:failure_cases}
  \end{center}
\end{figure}

Although \ACuRL{} achieves substantial improvements in both intra-environment and cross-environment continual learning settings, a significant gap remains in achieving reasonable performance, with success rates in some environments still below 50\%. To better understand these limitations, we conduct a detailed human analysis of our best-performing agent, which is continually trained across five environments to shed light on future development and improvement of CUAs.
We categorize agent failures into six mutually exclusive types based on root-cause analysis: (1) Knowledge \& Planning Limitations, where the agent lacks sufficient environmental knowledge of available UI elements and functionalities, or fails to formulate a coherent plan for compound multi-step tasks; (2) Premature Termination, characterized by the agent declaring success without verifying that the resulting state satisfies the task specification; (3) Visual Grounding Error, arising from misidentification or mislocation of target UI elements during interaction; (4) Repetitive Action Loops, where the agent repeatedly executes the same action without producing any state change; (5) Task Misunderstanding, involving misinterpretation of the natural-language instruction's semantics; and (6) Others, covering miscellaneous failures that do not fall into any of the above categories. As shown in Figure~\ref{fig:failure_cases}, the failure distribution is dominated by knowledge and planning limitations (37\%) and premature termination (30.4\%), which together account for 67.4\% of all observed failures. Visual grounding errors (11.9\%) and repetitive action loops (11.5\%) contribute comparably, with the latter highlighting the agent's inability to adaptively recover from unsuccessful actions. Task misunderstanding (7.0\%) and other miscellaneous causes (2.2\%) constitute a long tail of edge cases. Overall, these results indicate that the dominant bottlenecks lie in environment knowledge, task planning, and self-verification, rather than in low-level perception or instruction comprehension.
\section{Detailed results for each environment}
\label{appendix:detailed_results_for_main_figure}
\input{tables/main_table}

\input{tables/appendix_task_examples}
\clearpage

\section{Prompts}
\label{appendix:prompts}

\centering
\begin{strip}
\begin{tcolorbox}[colframe=yellow!50!black, colback=yellow!10!white, title=Curriculum Generator]
\small
You are a professional instructional designer specializing in software learning task design. 
Your objective is to generate exactly new \{total tasks\} high-quality, realistic learning tasks based on the previous feedback.\\

You will be provided with:\\
1. A record of random software exploration, including several actions and corresponding screenshots.\\
    - This shows the possible functionality of the software.\\
2. An initial state of the software (with its own screenshots).\\
    - This represents the exact starting point from which learners will begin.\\
3. A summary of learner performance on earlier tasks.\\  
    - This indicates which tasks were difficult, what the learner can already solve, and what gaps or weaknesses remain.\\

\#\# Software Exploration\\
Actions:
\{exploration actions\}\\

The corresponding \{len(screenshots)\} screenshots across the actions: \{screenshots\}\\

\#\# Initial State\\
The current initial state is shown in \{len(screenshots)\} additional screenshots. These screenshots show the starting point for the software.\\

The initial state screenshots: \{screenshots\}\\

\#\# Task Generation Requirements\\
Based on the random exploration actions and screenshots, combined with the initial state screenshots, generate exactly \{total tasks\} specific tasks that:

1. **High-Level User Goals (NO Step-by-Step Instructions Allowed)**\\  
    - Write tasks as **natural user goals**, NOT tutorials or procedures.\\  
    - Express tasks the way a real user would describe what they want to achieve.\\  
    - Absolutely **do NOT describe UI navigation paths** (e.g., “click X”, “open Y”, “go to View > Animation”).\\  
    - No operational verbs tied to UI mechanics (e.g., “open”, “select”, “navigate to”, “choose”).\\  
    - Use **intent-oriented phrasing**, not action-oriented phrasing.  
    - Each task must be **one or two concise sentences** describing the desired outcome or goal.\\

2. **Anchor in Initial State**\\
    - Each task must assume the learner starts from the provided initial state screenshots.\\  
    - Do not invent extra elements not visible in the initial state.\\  

3. **Leverage Exploration Evidence**\\  
    - Use the random exploration actions/screenshots to infer what the software can do.\\  
    - Design tasks that are plausible and valuable given those capabilities.\\  

4. **Realistic User Value**\\  
    - Tasks should reflect meaningful activities a real user of this software might want to accomplish, not artificial button-clicking exercises.\\    

5. **Distinct Coverage**\\  
    - Each of the task should focus on a different aspect, function, or workflow of the software.\\  
    - Avoid overlap. Aim for variety (e.g., editing, organizing, exporting, formatting).\\   

6. **Diversity**\\
    - Avoid creating tasks similar to the existing ones from previous feedback.\\
    - Generate creative variations and novel approaches.\\
    - Ensure variety in complexity, action types, and user scenarios.\\
    - Think of different user personas and use cases.\\
\end{tcolorbox}

\begin{tcolorbox}[colframe=yellow!50!black, colback=yellow!10!white, title=Curriculum Generator]
\small
7. **Clarity and Specificity (Anti-Ambiguity Rule)**\\  
    All regenerated tasks must be stated with full clarity and without ambiguity.\\  
    To ensure this:\\
    - Do not use vague or illustrative expressions such as “for example”, “such as”, “any”, “multiple”, or other open-ended terms.\\
    - Do not introduce optionality, alternatives, or flexible ranges (e.g., “A or B”, “choose whichever”, “a few”, “some”).\\
    - Do not reference hypothetical or representative instances.\\
    - Every task must define a **single, explicit target** and a **single, unambiguous goal**.\\
    - The task must be specific enough that a system can determine **exact completion criteria** without guessing. \\

8. **Feedback Integration**\\  
    You must generate a substantively different new task based on the agent’s SR.\\
    The new task must not be a minor revision, paraphrase, or parameter change of the original task.\\

    \#\#\#\# **If SR $>$ 70\% (High performance)**\\
    The agent performs the original task reliably. You must therefore generate a **fully new** task that is **significantly more challenging**, with **no structural, semantic, or phrasing resemblance** to the original.\\

    The new task must satisfy all of the following:\\

    **1. Fully New Scenario**\\
    - Use a **completely different real user motivation**, context, and type of content.\\
    - Do NOT share the original task's goal type, UI element category, action verbs, workflow pattern, or problem structure.\\

    **2. Higher Complexity**\\
    The new task must be **meaningfully more complex**, for example by:
    - involving **multiple related sub-goals**,\\ 
    - introducing **conditional or constraint-based goals**,\\ 
    - requiring **cross-region or cross-document coordination**, \\
    - adding **explicit constraints, requirements, or trade-offs**, \\
    - or requiring **richer interpretation or reasoning** about the interface/content.\\

    **3. Strict Non-Similarity**\\
    The new task must NOT:\\
    - resemble the original in structure or step pattern,\\
    - reuse similar keywords, verbs, UI components, or operations,\\
    - operate in the same functional space as the original (e.g., if original used formatting, avoid formatting),\\
    - pursue any version of the same underlying objective.\\

    The final task must feel like an **entirely new, high-value, realistic scenario** that a different user might genuinely perform.\\

\end{tcolorbox}

\begin{tcolorbox}[colframe=yellow!50!black, colback=yellow!10!white, title=]
\small
    \#\#\#\# **If 30\% $\le$ SR $\le$ 70\% (Medium performance)**\\
    The agent shows partial competence. You must generate a **different task of similar overall difficulty**, but with a **distinct scenario and distinct user intent**.\\

    You may:\\
    - explore a **new real user scenario**, or\\
    - create a **lateral variation** that practices similar reasoning skills without repeating the original pattern.\\

    Ensure meaningful variation by changing at least one of:\\
    - the **user persona**,\\
    - the **purpose or motivation**,\\
    - the **content being worked on**,\\
    - or the **overall context** (document type, audience, situation).\\

    The new task must **not** be a softened or strengthened rewrite of the original.\\

    \#\#\#\# **If SR $<$ 30\% (Low performance)**\\
    When the agent struggles significantly, generate a **scaffolded prerequisite task** that focuses on the fundamental skill(s) required to perform the original task. \\

    This scaffolded task must:\\
    - isolate **one essential underlying capability** needed to accomplish the original task,\\
    - be formulated as an **independent, authentic user goal**, not a simplified or partially completed version of the original task,\\
    - introduce a **distinct scenario, purpose, or context**, while still remaining within the same broad capability domain,\\
    - and explicitly target the foundational competencies the agent lacks (e.g., identifying relevant elements, applying targeted adjustments, interpreting context, selecting appropriate options, or manipulating core components).\\

    The new task should reflect a **core sub-skill** that the user must master before attempting any higher-level or more complex versions of the original task.\\

\{feedback section\}

Return the tasks in the following JSON structure:

```json
\{\{
    "tasks": [
        \{\{
            "task\_id": "task\_1",
            "original\_task\_description": "The original task description from the previous feedback",
            "reasoning": "Explain why and how this task was modified given the agent's SR and feedback context.",
            "new\_task\_description": "High-level yet specific user goal derived from the initial state and refined through feedback."
        \}\},\\
        \{\{
            "task\_id": "task\_2",
            "original\_task\_description": "The original task description from the previous feedback",
            "reasoning": "Explain why and how this task was modified given the agent's SR and feedback context.",
            "new\_task\_description": "High-level yet specific user goal derived from the initial state and refined through feedback."
        \}\},\\
        \{\{
            "task\_id": "task\_3",
            "original\_task\_description": "The original task description from the previous feedback",
            "reasoning": "Explain why and how this task was modified given the agent's SR and feedback context.",
            "new\_task\_description": "High-level yet specific user goal derived from the initial state and refined through feedback."
        \}\},\\
        \{\{
            "task\_id": "task\_4",    
            "original\_task\_description": "The original task description from the previous feedback",
            "reasoning": "Explain why and how this task was modified given the agent's SR and feedback context.",
            "new\_task\_description": "High-level yet specific user goal derived from the initial state and refined through feedback."
        \}\},\\
        ......
    ]
\}\}
'''
\end{tcolorbox}

\begin{tcolorbox}[colframe=yellow!50!black, colback=yellow!10!white, title=CUAJudge - Outcome Judgement]
\small

You are an expert evaluator of computer use agents. Your task is to rigorously assess whether an agent has successfully completed a given task by examining the initial state, final state, action history, and intermediate screenshots.\\

\#\# Evaluation Process\\

Follow this systematic evaluation process:\\

\#\#\# Step 1: Understand the Task Requirements\\
- Read the task description carefully and identify what needs to be accomplished\\
- Review the key points that define successful completion\\
- Examine the initial screenshot(s) to understand the starting state\\

\#\#\# Step 2: Compare Initial vs Final State\\
- **Critical**: Compare the initial screenshot(s) with the final screenshot side-by-side\\
- Identify what specific changes occurred in the interface\\
- Determine if these changes align with the task requirements\\
- Check if the final state demonstrates completion of all required modifications\\

\#\#\# Step 3: Verify Each Key Point\\
For EACH key point listed, systematically check:\\
- Is this requirement explicitly addressed in the action history?\\
- Is the result visible in the final screenshot or intermediate screenshots?\\
- Is the implementation correct (not just superficially similar)?
- Are the exact specifications met (correct values, correct targets, correct methods)?\\

\#\#\# Step 4: Apply Strict Evaluation Criteria\\
1: The agent must only make changes that are directly necessary to complete the task. Any unrelated edits or system/software modifications that are not explicitly required for task completion will result in failure.\\
2: Interpret the user's instructions in the context of the current software. Professional or common software terms have precise meanings in their respective applications. An action that superficially matches the wording but targets the wrong function or interface region of the software should be considered a failure.\\
3: You must carefully check whether these snapshots and action history meet these key points. Ensure that specific filter conditions, such as "best," "highest," "cheapest," "latest," "most recent," "lowest," "closest," "highest-rated," "largest," and "newest" are correctly applied using the filter function (e.g., sort function).\\
4: If the agent loops through a sequence of actions that do not make progress toward the goal (including failing to click "Save" or "Submit," etc.), it should be considered a failure.\\

\#\#\# Step 5: Make Final Judgment\\

**SUCCESS** requires:\\
- ALL key points are satisfied\\
- ALL evaluation criteria are met\\
- Final screenshot shows correct end state\\
- No extraneous or incorrect modifications\\

**FAILURE** if ANY of these are true:\\
- Even one key point is not satisfied\\
- Any evaluation criterion is violated\\
- Final state does not match requirements\\
- Evidence of incorrect or incomplete execution

\end{tcolorbox}

\begin{tcolorbox}[colframe=yellow!50!black, colback=yellow!10!white, title=]
\small
\#\# Response Format\\

Provide your evaluation in exactly two lines:\\

Thoughts: [Systematically walk through Steps 1-5 above. For each key point, explicitly state whether it's satisfied and cite evidence from screenshots/actions. Compare initial and final states. Note any violations of criteria.]\\
Status: "success" or "failure"\\

**Important**: Be thorough but concise. Your thoughts should demonstrate you checked each key point and criterion.\\

\#\# Task to Evaluate: \{task\}\\

\#\# Key Points for Successful Completion: \{key points\}\\

\#\# Agent's Action History: \{action history\}\\

\end{tcolorbox}
\end{strip}

\clearpage
\section{Examples of Context Across Different Environments}
\label{appendix:example_contexts}

\begin{figure*}[h]
  \centering
  \begin{subfigure}[h]{0.48\textwidth}
    \centering
    \includegraphics[width=\linewidth]{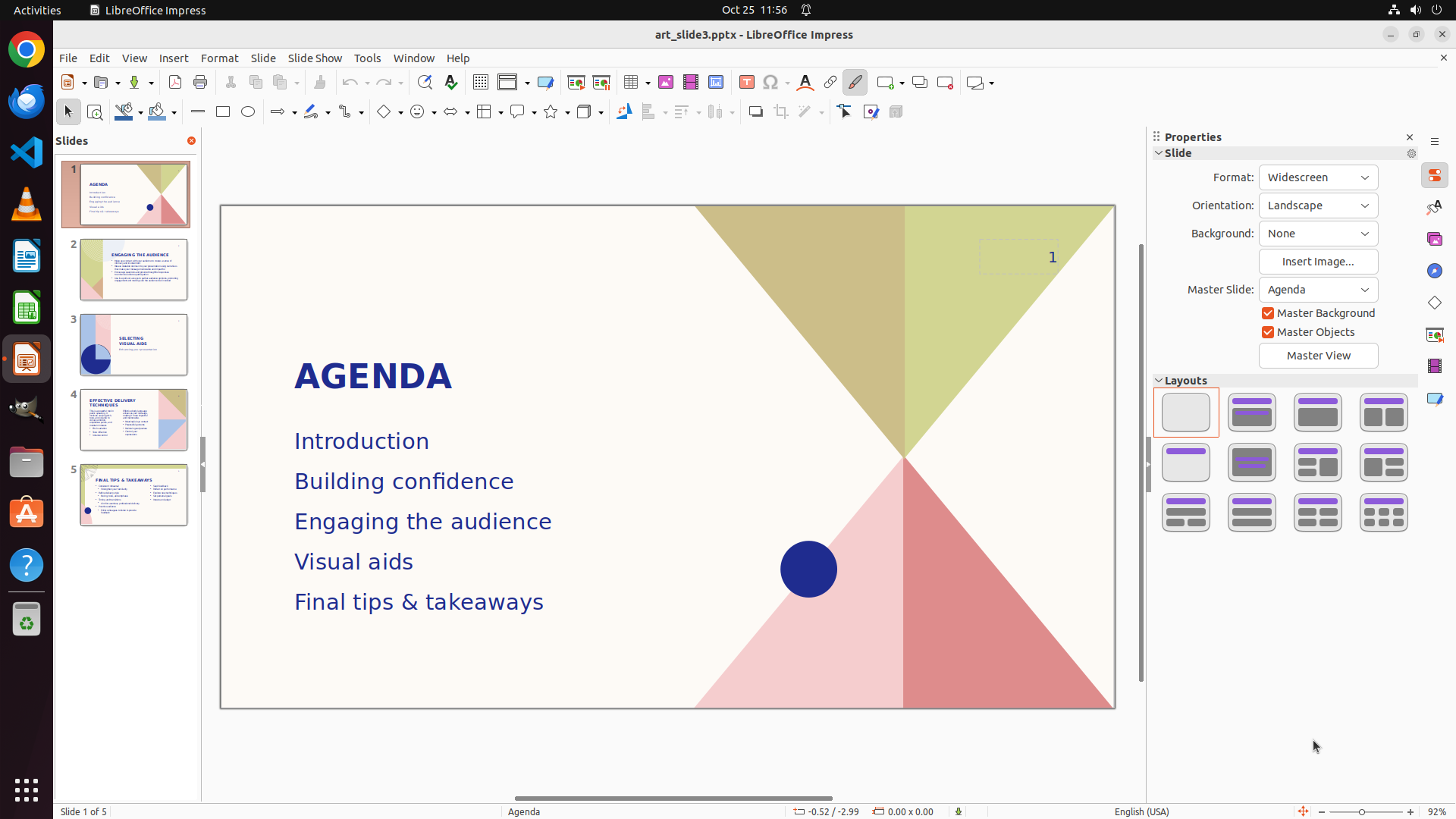}
    \caption{Example 1}
  \end{subfigure}
  \hfill
  \begin{subfigure}[h]{0.48\textwidth}
    \centering
    \includegraphics[width=\linewidth]{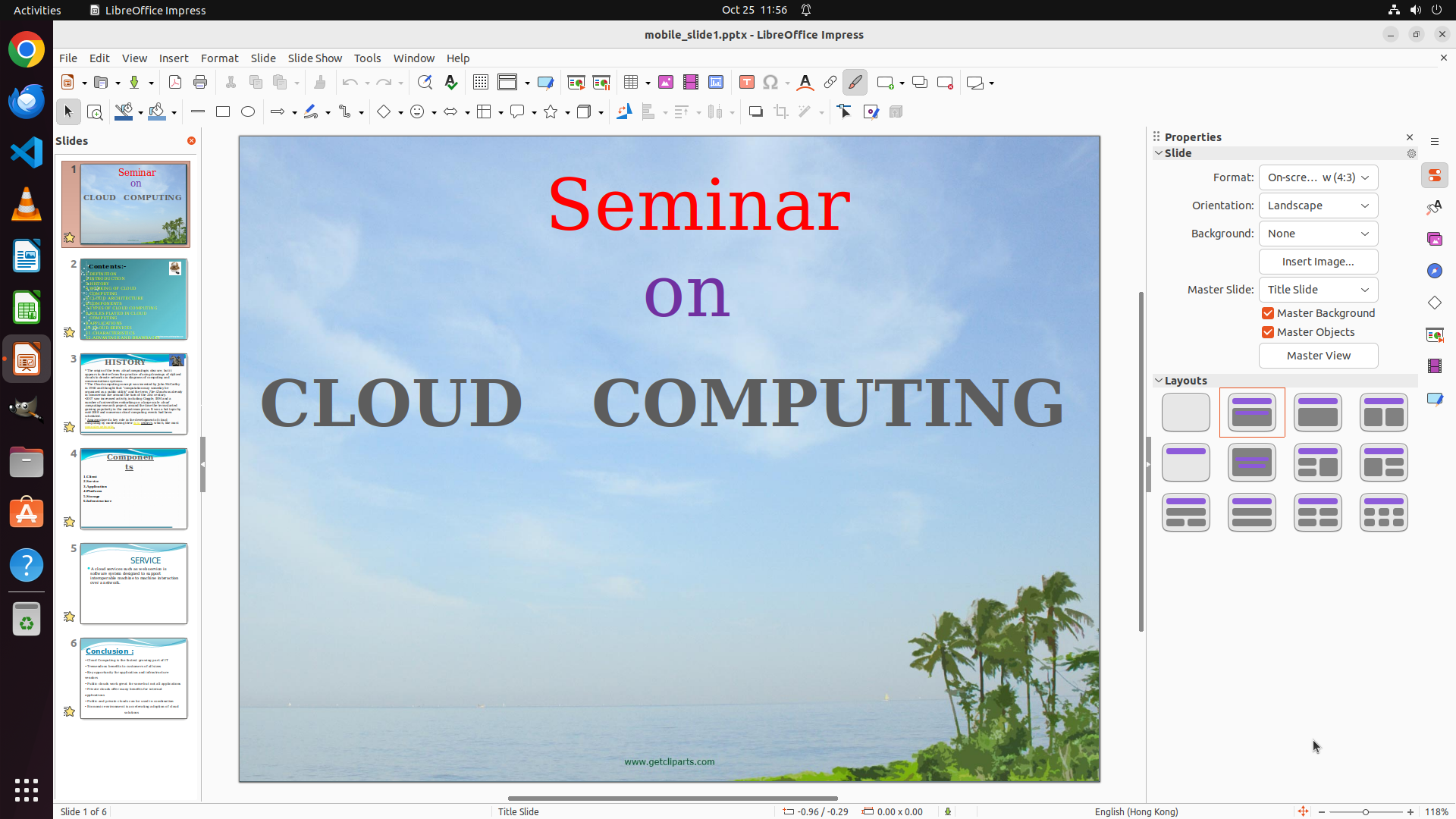}
    \caption{Example 2}
  \end{subfigure}
  \caption{Examples of LibreOffice Impress context.}
\end{figure*}

\begin{figure*}[h]
  \centering
  \begin{subfigure}[h]{0.48\textwidth}
    \centering
    \includegraphics[width=\linewidth]{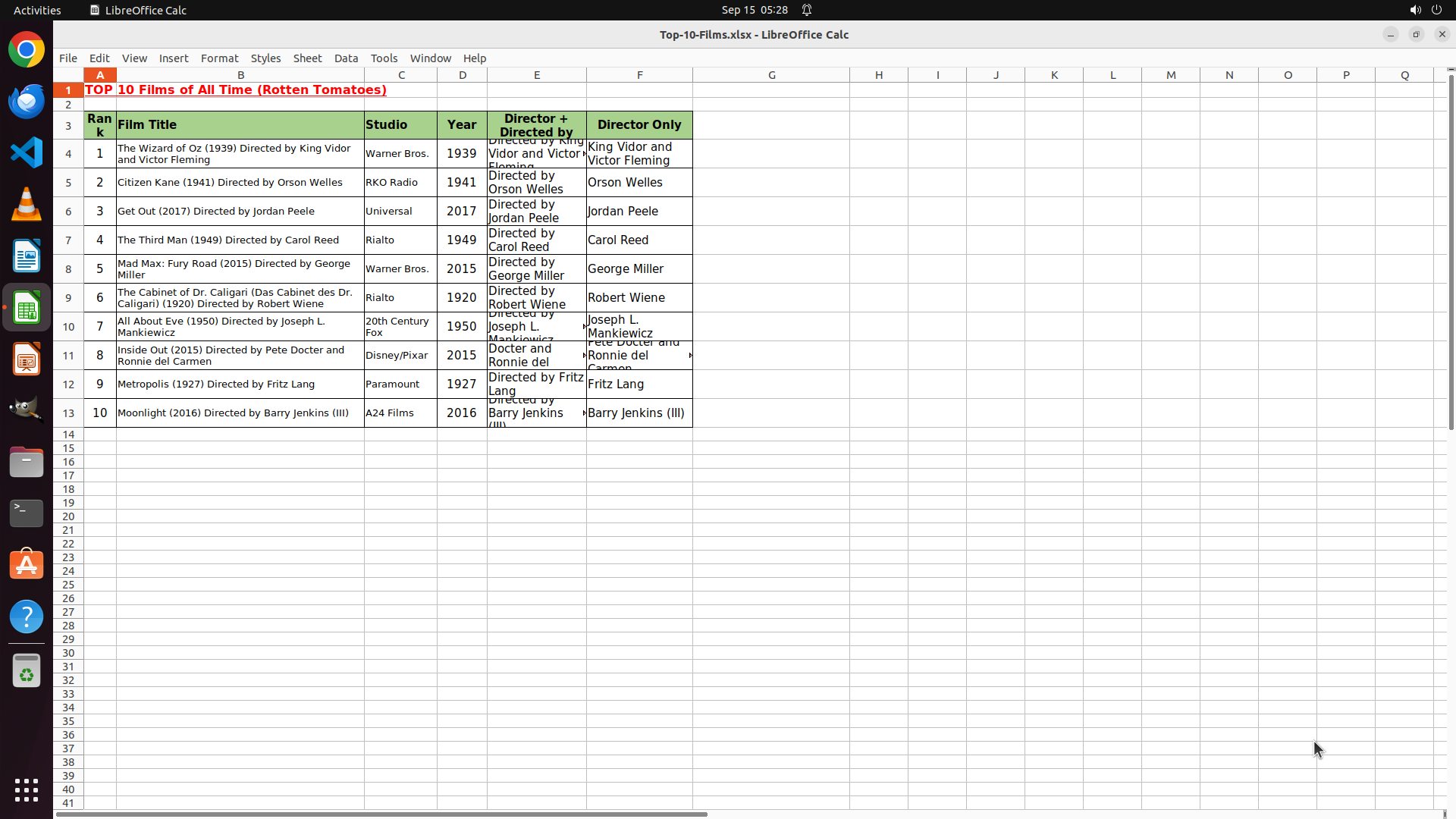}
    \caption{Example 1}
  \end{subfigure}
  \hfill
  \begin{subfigure}[h]{0.48\textwidth}
    \centering
    \includegraphics[width=\linewidth]{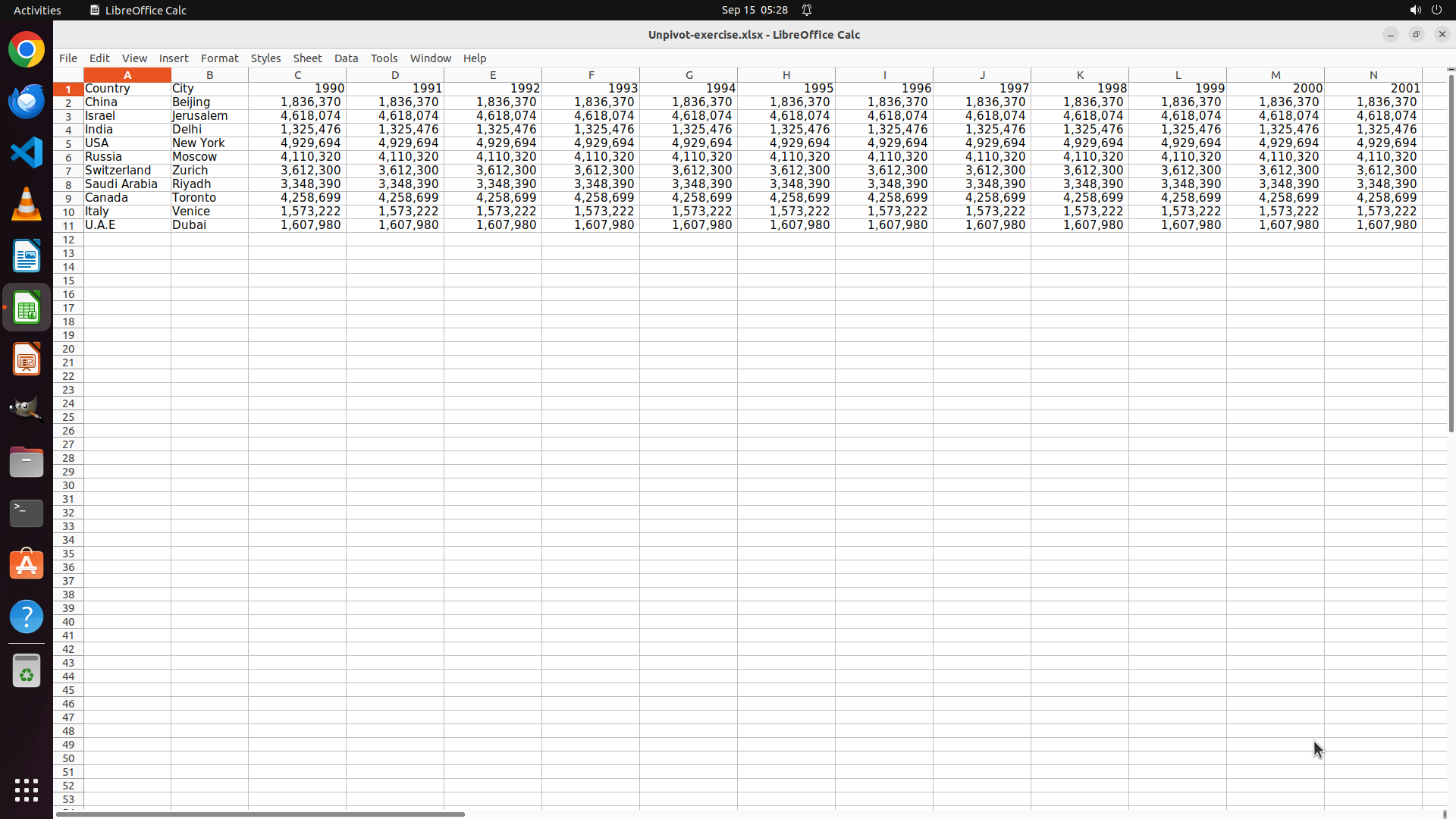}
    \caption{Example 2}
  \end{subfigure}
  \caption{Examples of LibreOffice Calc context.}
\end{figure*}

%% file: sections/related_work.tex
\section{Related Work}
\label{appendix:related_work}
\subsection{GUI Agent}
GUI agents operate directly on digital environments to accomplish user-specified tasks by integrating perception~\citep{gou2025navigating}, planning~\citep{Gu2025WebDreamer}, and reasoning~\citep{qin2025ui}. Existing work has explored a wide range of interaction paradigms, spanning from DOM-based web manipulation~\citep{zhou2024webarena,deng2023mind2web,wei2025webagent,qi2025webrl} to pixel-level control in desktop environments~\citep{anthropic2024computeruse,anthropic2025computeruse,qin2025ui,liao2025beyond}. 
Building on these paradigms, prior studies have significantly advanced the capabilities of GUI agents. However, they either heavily rely on large-scale human-annotated data~\citep{wang2025opencuaopenfoundationscomputeruse,liu2025scalecua,deng2023mind2web,wang2025uitars2technicalreportadvancing,feizi2025grounding,he2025efficient}, which is costly and unscalable, or synthesize data from a limited set of environments~\citep{sun2025seagent,qi2025webrl,wei2025webagent,yang2025ultracua,awadallah2025fara}, failing to capture the diversity of millions of desktop and web applications in the real-world scenarios. Moreover, they model the digital world as static, despite its inherently dynamic nature. In practice, interfaces and functionalities continually evolve~\citep{pan2024webcanvas,xue2025an}, which can lead to substantial performance degradation when such changes are not captured by prior training data~\citep{mirzadeh2025gsmsymbolic,cuvin2025decepticon}. 
Although~\citet{liu2026continual} studies real-world CUA scenarios where data is naturally in flux, its focus is primarily on grounding rather than end-to-end CUA task execution. Consequently, it does not capture the planning and multi-step decision-making required to solve realistic CUA tasks, which is a key distinction from our work. Moreover, they consider only two scenarios: platform migration (mobile, desktop, and web) and resolution changes. In contrast, our work addresses a broader and more realistic range of dynamics, including cross-environment continual learning scenarios and environment updates such as software version upgrades.
Overall, we move beyond the static environment assumption and explicitly model real-world digital environments as dynamic and evolving, better reflecting practical deployment scenarios for end-to-end computer-use agents.

\subsection{Autonomous Learning without Human Data}
To reduce reliance on human effort, reinforcement learning~\citep{ouyang2022training,schulman2017proximal} has emerged as a popular paradigm, as it enables agents to learn from environment interactions via reward feedback without explicit human supervision~\citep{silver2016mastering,silver2018general,schrittwieser2020mastering}.
While existing approaches show promise, transferring them to the computer-use domain remains challenging, as they target weakly grounded or environment-agnostic tasks that rely primarily on internal knowledge for synthesis.~\citep{liu2025agent0,huang2025r,zhao2025absolute,wei2025toward,liu2025spiralselfplayzerosumgames,fang2025serl}. In contrast, computer-use tasks must be grounded in interactive environments. Such environment-specific experiences are largely absent from pretraining data, making it difficult to synthesize high-quality tasks based on the model's internal knowledge alone.
Moreover, these domains can easily obtain verifiable rewards, such as exact answer matching~\citep{cobbe2021training,kyidlicecek2025mathverify} and deterministic program execution~\citep{zhao2025absolute,wei2025toward,wei2025swe}; even open-ended instruction-following tasks can be alleviated via large-scale preference data~\citep{ouyang2022training,bai2022training,bai2022constitutional}.
By contrast, obtaining reliable reward signals in computer-use tasks is substantially more challenging~\citep{zhang2025agent}, as task completion cannot be inferred from the final state alone, and trajectories typically span long horizons with rich multimodal observations (e.g., hundreds of screenshots), rendering trajectory evaluation itself a nontrivial problem~\citep{xue2025an}. 
Although SEAgent~\citep{sun2025seagent} explores autonomous learning for computer-use agents, it is developed under a simplified setting that deviates from real-world requirements. First, it does not account for environment dynamics, such as software updates, platform migration, and resolution changes, which introduce substantial distribution shifts in real-world interfaces. Second, it focuses on specialist agents within a single environment, overlooking cross-environment continual learning. In practice, both environments and user needs (e.g., new environments and increasingly complex goals) evolve over time, making cross-environment continual learning essential.
Third, it adopts single-turn RL rather than end-to-end, multi-step RL. As a result, it fails to capture the emergent self-refinement and exploration behaviors that are essential for computer-use agents.
In this work, we systematically study continual learning for CUAs, bridging the gap to diverse and dynamic digital environments in the real world, where agents continually learn in both intra-environment and cross-environment scenarios, while adapting to environment shifts (e.g., software updates, platform migration, and resolution changes) through a fully autonomous, end-to-end curriculum RL framework.

%% file: sections/limitations.tex
\section{Limitations}
\label{appendix:limitations}
Although \ACuRL{} achieves substantial gains in both intra- and cross-environment continual learning autonomously without human supervision, it still depends on some proprietary models (i.e., GPT-5-mini for evaluation and GPT-5 for task generation). While replacing these components with the open-source Qwen3-VL-8B also yields modest improvements, \ACuRL{} still fails to enable fully autonomous self-evolution for agents with weaker instruction-following capabilities, such as UI-TARS. We leave this for future work, jointly enhancing GUI navigation and instruction-following, enabling agents to better support task generation and achieve true self-evolution.

%% file: sections/impact_statements.tex
\section{Impact Statement}
\label{appendix:impact_statements}
Autonomous computer-use agents have the potential to enhance productivity by assisting users in complex digital environments, such as office software and specialized applications. In this work, we study how such agents can continually learn and adapt to specific target environments without relying on human-annotated data. All experiments are conducted in controlled settings explicitly designed to prevent harmful actions on real-world systems.
At the same time, computer-use agents are capable of executing multi-step actions that modify environment states, which may lead to unintended or potentially harmful behaviors if deployed without adequate safeguards, such as accidentally modifying or deleting user files. These risks raise important safety and ethical considerations, particularly for real-world deployment. We therefore believe that future research should further explore safety-aware evaluation protocols and effective risk mitigation mechanisms. Our code and models are released solely for research purposes, and we strongly discourage any use that could harm users or violate privacy.

%% file: sections/background.tex
\section{Background}
\label{appendix:background}
We model a computer-use agent task execution as a Partially Observable Markov Decision Process (POMDP), where $\mathcal{S}$ denotes the environment state space, $\mathcal{O}$ the observation space, $\mathcal{A}$ the executable action space, $\mathcal{T}\colon \mathcal{S} \times \mathcal{A} \rightarrow \mathcal{S}$ the environment transition function, and $\mathcal{R}$ the reward function. At step $t$, the agent receives a partial observation $o_t \in \mathcal{O}$, which may consist of a natural language task description together with observations such as screenshots. Based on the interaction history $h_t = (o_{\le t}, a_{<t})$, the agent generates an executable action $a_t \in \mathcal{A}$, e.g., clicking a specific screen location or issuing a keyboard command, which transitions the environment to a new state $s_{t+1} \in \mathcal{S}$ and yields a new partial observation $o_{t+1} \in \mathcal{O}$. The interaction loop continues until the agent produces a termination action or reaches a predefined maximum number of steps. 
Finally, task completion is determined by an evaluator, who assigns an outcome $\mathcal{R}(\tau) \in [0,1]$ based on the entire interaction trajectory $\tau$. A value of $\mathcal{R}(\tau)=1$ indicates successful task completion, while $\mathcal{R}(\tau)=0$ indicates failure.

%% file: tables/geneator_comparison.tex
\begin{table}[h]
\centering
\small
\caption{Performance comparison of different generators on Impress at iteration 1, which involves curriculum learning from iteration 0 to 1.}
\label{tab:generator_impress}
\begin{tabular}{l c}
\toprule
\textbf{Generator} & \textbf{Impress} \\
\midrule
GPT-5 & \textbf{30.8} \\
Qwen3-VL-8B & 29.5 \\
\bottomrule
\end{tabular}
\end{table}

%% file: tables/CUAjudge_key_screenshot_identification_trade_off.tex
\begin{table}[h]
\centering
\caption{Comparison of precision and cost across GPT-5-mini and Qwen3-VL-8B models for the key screenshot identification module.}
\begin{tabular}{lcc}
\toprule
Model & Precision & Cost (\$) \\
\midrule
Qwen3-VL-8B-Instruct & \textbf{70.3} & \textbf{0.4} \\
GPT-5-mini           & 65.6          & 2.0          \\
\bottomrule
\end{tabular}
\label{tab:model_cost_precision}
\end{table}

%% file: tables/judge_ablation.tex
\begin{table*}[ht!]
\centering
\small
\caption{\textbf{W/o state difference} indicates that no state difference analysis is performed during the outcome judgment stage, thereby ablating this module.}
\resizebox{\textwidth}{!}{
\begin{tabular}{l|ccc|ccc|ccc|ccc|c}
\toprule
 & \multicolumn{3}{c|}{\textbf{Claude-Sonnet-4}}
 & \multicolumn{3}{c|}{\textbf{Claude-Sonnet-4.5}}
 & \multicolumn{3}{c|}{\textbf{O3}}
 & \multicolumn{3}{c|}{\textbf{OpenCUA-7B}}
 & \textbf{Overall} \\
\midrule
\textbf{Evaluator}
 & Precision & Recall & Agreement
 & Precision & Recall & Agreement
 & Precision & Recall & Agreement
 & Precision & Recall & Agreement
 & Agreement \\
\midrule
CUAJudge
 & \textbf{74.8} & \textbf{86.4} & \textbf{87.0}
 & 82.1 & \textbf{84.8} & \textbf{85.9}
 & 42.1 & 50.0 & 89.4
 & 66.2 & 71.0 & 87.5
 & \textbf{87.5} \\
\hspace{2mm} - w/o state difference
 & 70.5 & 76.1 & 82.5
 & \textbf{83.7} & 76.1 & 83.4
 & \textbf{44.4} & 48.5 & \textbf{89.7}
 & \textbf{72.1} & 68.1 & \textbf{88.4}
 & 86.0 \\
\bottomrule
\end{tabular}
}
\label{tab:judge_comparison_ablation}
\end{table*}

%% file: tables/task_statistics.tex
\begin{table}[h!]
\centering
\small
\caption{Task statistics for each environment.}
\label{tab:task_source}
\resizebox{\columnwidth}{!}{
\begin{tabular}{lcc}
\toprule
\textbf{Environment} & \textbf{Task Source} & \textbf{\#Tasks} \\
\midrule
LibreOffice Impress & OSWorld  & 47 \\
                    & OfficeWorld   & 60 \\
\midrule
LibreOffice Calc    & OSWorld  & 47 \\
                    & OfficeWorld   & 60 \\
\midrule
LibreOffice Writer  & OSWorld  & 23 \\
                    & OfficeWorld   & 60 \\
\midrule
Thunderbird         & OSWorld    & 15 \\
\midrule
Celestia            & ScienceBoard & 33 \\
\midrule
KAlgebra            & ScienceBoard & 31 \\
\bottomrule
\end{tabular}
\label{tab:task_statistics}
}
\end{table}

%% file: tables/comparison_baselines.tex
\begin{table}[h!]
\centering
\small
\caption{
Performance comparison with curriculum-based baselines on OSWorld and ScienceBoard, including DigiRL~\citep{bai2024digirltraininginthewilddevicecontrol}, WebRL~\citep{qi2025webrl}, and SEAgent~\citep{sun2025seagent}.
}
\label{tab:main_comparison}
\begin{tabular}{lcccc}
\toprule
\textbf{Method} & \textbf{Impress} & \textbf{Writer} & \textbf{KAlgebra} & \textbf{Celestia} \\
\midrule
DigiRL & 19.6 & 19.1 & - & - \\
WebRL & 20.4 & 21.7 & - & - \\
SEAgent & 31.9 & \textbf{43.5} & \textbf{29} & 15.2 \\
ACuRL & \textbf{44.5} & \textbf{43.5} & 28 & \textbf{24.2} \\
\bottomrule
\end{tabular}
\end{table}

%% file: tables/main_table.tex
\begin{table*}[h!]
\centering
\small
\caption{
Average success rate (± standard deviation) across six environments for the base agent and agents that continually learn in specific environments via \ACuRL{}. Rows correspond to agents continually learning in a specific environment and columns to testing environments. The overall score is the mean performance over all six environments. \textbf{Bold} denotes the best performance.}
\resizebox{\textwidth}{!}{
\begin{tabular}{lccccccc|c}
\toprule
\textbf{Agent} & \textbf{Iteration}
& \textbf{Impress}
& \textbf{Calc}
& \textbf{Writer}
& \textbf{Thunderbird}
& \textbf{KAlgebra}
& \textbf{Celestia}
& \textbf{Overall} \\
\midrule

UI-TARS-1.5-7B (Base agent)
& --
& 31.1\std{1.0}
& 6.5\std{1.6}
& 15.7\std{2.0}
& 35.5\std{3.9}
& 10.8\std{1.8}
& 17.2\std{1.7}
& 19.5\std{0.9} \\
\midrule

\multirow{3}{*}{Impress}
& Iter 1
& \cellcolor{diagcolor}34.2\std{0.5}
& 5.9\std{0.6}
& 18.9\std{0.7}
& 44.5\std{3.9}
& 14.0\std{1.8}
& 19.2\std{1.7}
& 22.8\std{0.6} \\
& Iter 2
& \cellcolor{diagcolor}38.8\std{0.6}
& 8.1\std{0.5}
& 21.7\std{0.0}
& 51.1\std{3.8}
& 14.0\std{1.8}
& 20.2\std{1.7}
& \mean{25.7}\std{0.3} \\
& Iter 3
& \cellcolor{diagcolor}\mean{40.7}\std{1.0}
& 5.9\std{1.1}
& 20.9\std{1.4}
& 40.0\std{6.7}
& 12.9\std{0.0}
& 23.2\std{4.6}
& 23.9\std{0.8} \\
\midrule

\multirow{3}{*}{Calc}
& Iter 1
& 32.3\std{1.4}
& \cellcolor{diagcolor}7.5\std{0.0}
& 19.3\std{1.3}
& 44.5\std{3.9}
& 12.9\std{3.2}
& 16.2\std{1.7}
& 22.1\std{0.3} \\
& Iter 2
& 31.7\std{1.0}
& \cellcolor{diagcolor}7.8\std{1.1}
& 18.5\std{0.7}
& 46.7\std{0.0}
& 19.4\std{3.3}
& 13.2\std{3.5}
& 22.9\std{1.2} \\
& Iter 3
& 33.2\std{2.4}
& \cellcolor{diagcolor}\mean{9.0}\std{0.5}
& 16.9\std{1.3}
& 37.8\std{3.9}
& 16.1\std{3.3}
& 13.1\std{1.8}
& 21.0\std{0.6} \\
\midrule

\multirow{3}{*}{Writer}
& Iter 1
& 30.7\std{2.5}
& 6.9\std{1.1}
& \cellcolor{diagcolor}18.9\std{1.8}
& 44.5\std{3.9}
& 14.0\std{1.8}
& 17.2\std{1.7}
& 22.0\std{0.3} \\
& Iter 2
& 32.9\std{0.5}
& 5.6\std{0.9}
& \cellcolor{diagcolor}20.5\std{1.2}
& 51.1\std{3.8}
& 10.8\std{1.8}
& 17.2\std{4.6}
& 23.0\std{1.5} \\
& Iter 3
& 30.8\std{1.9}
& 7.4\std{1.6}
& \cellcolor{diagcolor}\mean{22.5}\std{1.8}
& 44.4\std{7.7}
& 17.2\std{4.9}
& 17.2\std{1.7}
& 23.3\std{0.7} \\
\midrule

\multirow{3}{*}{Thunderbird}
& Iter 1
& 31.1\std{1.9}
& 6.6\std{1.0}
& 17.7\std{0.7}
& \cellcolor{diagcolor}42.2\std{3.9}
& 10.8\std{3.7}
& 16.2\std{1.7}
& 20.8\std{1.6} \\
& Iter 2
& 35.4\std{1.0}
& 5.6\std{0.0}
& 18.1\std{4.4}
& \cellcolor{diagcolor}42.2\std{3.9}
& 8.6\std{4.9}
& 20.2\std{4.6}
& 21.7\std{1.7} \\
& Iter 3
& 35.1\std{1.9}
& 7.5\std{1.0}
& 13.7\std{0.7}
& \cellcolor{diagcolor}\mean{57.8}\std{3.9}
& 14.0\std{3.7}
& 22.2\std{4.6}
& 25.1\std{0.4} \\
\midrule

\multirow{3}{*}{KAlgebra}
& Iter 1
& 32.3\std{2.8}
& 6.5\std{2.8}
& 19.3\std{2.1}
& 51.1\std{3.8}
& \cellcolor{diagcolor}18.3\std{1.9}
& 22.2\std{6.2}
& 25.0\std{1.4} \\
& Iter 2
& 29.8\std{2.5}
& 7.1\std{1.1}
& 16.8\std{1.3}
& 46.7\std{0.0}
& \cellcolor{diagcolor}20.5\std{1.8}
& 19.2\std{4.6}
& 23.4\std{0.4} \\
& Iter 3
& 30.8\std{1.6}
& 5.3\std{1.0}
& 16.9\std{2.5}
& 48.9\std{3.8}
& \cellcolor{diagcolor}\mean{28.0}\std{4.9}
& 14.2\std{1.8}
& 24.0\std{1.0} \\
\midrule

\multirow{3}{*}{Celestia}
& Iter 1
& 33.6\std{2.4}
& 5.6\std{0.0}
& 22.1\std{0.7}
& 48.9\std{3.8}
& 11.8\std{1.8}
& \cellcolor{diagcolor}22.2\std{1.7}
& 24.0\std{0.2} \\
& Iter 2
& 31.7\std{1.6}
& 5.0\std{1.4}
& 22.1\std{1.8}
& 42.2\std{3.9}
& 15.0\std{1.8}
& \cellcolor{diagcolor}23.2\std{1.7}
& 23.2\std{0.3} \\
& Iter 3
& 34.2\std{1.6}
& 7.5\std{1.0}
& 22.1\std{0.7}
& 48.9\std{3.8}
& 15.0\std{1.8}
& \cellcolor{diagcolor}\mean{24.2}\std{3.1}
& 25.3\std{0.4} \\

\midrule
Qwen3-VL-8B (Base agent)
& --
& 27.0\std{1.6}
& 15.3\std{1.9}
& 18.9\std{2.5}
& 51.1\std{3.8}
& 9.7\std{3.2}
& 10.1\std{1.7}
& 22.0\std{0.7} \\
\midrule

\multirow{3}{*}{Impress}
& Iter 1
& \cellcolor{diagcolor}30.8\std{2.5}
& 14.4\std{1.1}
& 21.3\std{0.7}
& 66.7\std{0.0}
& 11.8\std{1.8}
& 6.1\std{0.0}
& 25.2\std{0.6} \\
& Iter 2
& \cellcolor{diagcolor}36.0\std{1.1}
& 14.0\std{0.0}
& 22.9\std{1.2}
& 68.9\std{3.8}
& 17.2\std{1.9}
& 11.1\std{1.7}
& 28.4\std{0.7} \\
& Iter 3
& \cellcolor{diagcolor}\mean{39.2}\std{1.0}
& 14.0\std{1.9}
& 24.9\std{1.8}
& 64.5\std{3.9}
& 17.2\std{1.9}
& 11.1\std{1.7}
& 28.5\std{1.2} \\
\midrule

\multirow{3}{*}{Calc}
& Iter 1
& 25.2\std{1.6}
& \cellcolor{diagcolor}16.2\std{0.5}
& 20.5\std{1.3}
& 66.7\std{0.0}
& 10.8\std{1.8}
& 10.1\std{1.7}
& 24.9\std{0.7} \\
& Iter 2
& 27.3\std{1.1}
& \cellcolor{diagcolor}19.5\std{2.7}
& 22.5\std{0.7}
& 66.7\std{0.0}
& 11.8\std{1.8}
& 13.1\std{1.8}
& 26.8\std{0.5} \\
& Iter 3
& 33.0\std{1.9}
& \cellcolor{diagcolor}\mean{24.2}\std{2.5}
& 26.5\std{1.2}
& 62.2\std{3.9}
& 10.8\std{1.8}
& 11.1\std{1.7}
& 28.0\std{0.4} \\
\midrule

\multirow{3}{*}{Writer}
& Iter 1
& 26.4\std{0.5}
& 13.4\std{0.5}
& \cellcolor{diagcolor}23.3\std{0.7}
& 68.9\std{3.8}
& 10.8\std{1.8}
& 10.1\std{1.7}
& 25.5\std{1.3} \\
& Iter 2
& 26.1\std{1.9}
& 14.4\std{1.1}
& \cellcolor{diagcolor}29.7\std{1.8}
& 71.1\std{3.8}
& 18.3\std{1.9}
& 11.1\std{1.7}
& 28.5\std{0.8} \\
& Iter 3
& 27.7\std{0.5}
& 17.1\std{0.6}
& \cellcolor{diagcolor}\mean{30.1}\std{3.2}
& 68.9\std{3.8}
& 12.9\std{3.2}
& 15.2\std{3.1}
& 28.7\std{1.0} \\
\midrule

\multirow{3}{*}{Thunderbird}
& Iter 1
& 27.0\std{1.0}
& 12.5\std{1.1}
& 22.9\std{2.4}
& \cellcolor{diagcolor}66.7\std{0.0}
& 9.7\std{0.0}
& 11.1\std{1.7}
& 25.0\std{0.3} \\
& Iter 2
& 29.5\std{1.4}
& 9.4\std{1.0}
& 20.1\std{1.4}
& \cellcolor{diagcolor}71.4\std{3.2}
& 11.8\std{1.8}
& 8.1\std{1.7}
& 25.1\std{0.7} \\
& Iter 3
& 33.0\std{1.1}
& 11.8\std{1.1}
& 24.1\std{1.2}
& \cellcolor{diagcolor}\mean{80.0}\std{0.0}
& 12.9\std{3.2}
& 11.1\std{1.7}
& \textbf{28.8}\std{0.4} \\
\midrule

\multirow{3}{*}{KAlgebra}
& Iter 1
& 27.3\std{4.4}
& 19.3\std{2.0}
& 21.7\std{1.2}
& 42.2\std{3.9}
& \cellcolor{diagcolor}11.8\std{1.8}
& 6.1\std{0.0}
& 21.4\std{0.3} \\
& Iter 2
& 30.4\std{1.9}
& 15.3\std{0.5}
& 25.3\std{1.2}
& 55.5\std{3.9}
& \cellcolor{diagcolor}20.5\std{1.8}
& 10.1\std{1.7}
& 26.2\std{0.5} \\
& Iter 3
& 30.1\std{2.1}
& 17.1\std{1.4}
& 28.5\std{0.7}
& 62.2\std{3.9}
& \cellcolor{diagcolor}\mean{26.9}\std{1.8}
& 8.1\std{1.7}
& \textbf{28.8}\std{0.6} \\
\midrule

\multirow{3}{*}{Celestia}
& Iter 1
& 25.4\std{0.6}
& 14.6\std{1.9}
& 19.3\std{1.2}
& 60.0\std{6.7}
& 9.7\std{3.2}
& \cellcolor{diagcolor}13.5\std{2.4}
& 23.8\std{0.9} \\
& Iter 2
& 27.0\std{1.0}
& 14.3\std{1.1}
& 20.9\std{0.7}
& 58.1\std{8.3}
& 10.8\std{1.8}
& \cellcolor{diagcolor}17.2\std{1.7}
& 24.7\std{1.8} \\
& Iter 3
& 26.4\std{0.5}
& 12.2\std{1.0}
& 18.1\std{0.0}
& 62.2\std{3.9}
& 12.9\std{3.2}
& \cellcolor{diagcolor}\mean{18.2}\std{3.0}
& 25.0\std{1.6} \\

\bottomrule
\end{tabular}
}
\label{tab:specialist_models}
\end{table*}

%% file: tables/appendix_task_examples.tex
\begin{table*}[!t]
    \centering
    \small
    \caption{Task examples from different environments across 3 iterations.}
    \resizebox{\textwidth}{!}{
    \begin{tabular}{cp{13cm}}
    \toprule
        \textbf{Iteration} & \textbf{Task Description} \\
        \midrule
        \multicolumn{2}{c}{\textbf{LibreOffice Impress}} \\
        \midrule
        1 & Open Slide Properties for slide 5 and change the slide name to ``Future Trends". \\
        2 & On slide 4, append a numbered list with exactly these items: `Smart home devices', `Personalized shopping', `Healthcare services'. \\
        3 & On slide 5, add a straight connector from the apex of the pyramid to the text ``Human-AI Collaboration". \\
    \midrule
        \multicolumn{2}{c}{\textbf{LibreOffice Writer}} \\
        \midrule
        1 & Insert a filled rectangle below the body text and type ``Scan here" inside it. \\
        2 & Present the document in two columns while keeping the shaded block at the top as a full‑width section above the columns. \\
        3 & After the line “How to learn this vocabulary”, insert a cross-reference that displays the page number where “Graduate” appears. \\
    \midrule
        \multicolumn{2}{c}{\textbf{LibreOffice Calc}} \\
        \midrule
        1 & Apply a light gray background fill to the header cells A1:F1. \\
        2 & Create a filtered copy of all rows where Transmission (column E) equals ``Manual" into a new sheet named Manual Cars. \\
        3 & Sort the table so Areas follow the custom order North, Midlands, Scotland, South-East, South-West while preserving the row integrity of each record. \\
    \midrule
        \multicolumn{2}{c}{\textbf{Thunderbird}} \\
        \midrule
        1 & Create a local folder named Reports 2025 and move both messages from the Bills folder into it. \\
        2 & Insert the clickable mail link mailto:assistant@outlook.com on its own line in the message body. \\
        3 & Create a Saved Search folder named ``Recent Attachments" that lists messages with attachments from Inbox and Local Folders $>$ Bills received within the last 60 days, and ensure this saved search appears in Favorites. \\
    \midrule
        \multicolumn{2}{c}{\textbf{KAlgebra}} \\
        \midrule
        1 & Compute $sin(0.5)^2$ and report the numeric value. \\
        2 & Visualize the 3D surface $z = x^2 + y^2$ on the 3D graph. \\
        3 & Show two 3D surfaces simultaneously: $z = x^2 - y^2$ and $z = sin(x) + sin(y)$. \\
    \midrule
        \multicolumn{2}{c}{\textbf{Celestia}} \\
        \midrule
        1 & Turn off the Deep Sky Objects layer so only stars and planets remain visible. \\
        2 & Jump to 2020‑01‑01 00:00 in local time, enable light‑travel delay, and keep the simulation paused at that moment. \\
        3 & Display only the Galactic grid while keeping the Equatorial, Ecliptic, Horizontal grids and the ecliptic line hidden. \\
    \bottomrule
    \end{tabular}}
    \label{tab:task_example}
\end{table*}

%% file: reference.bib
@inproceedings{
    yu2025dapo,
    title={{DAPO}: An Open-Source {LLM} Reinforcement Learning System at Scale},
    author={Qiying Yu and Zheng Zhang and Ruofei Zhu and Yufeng Yuan and Xiaochen Zuo and YuYue and Weinan Dai and Tiantian Fan and Gaohong Liu and Juncai Liu and LingJun Liu and Xin Liu and Haibin Lin and Zhiqi Lin and Bole Ma and Guangming Sheng and Yuxuan Tong and Chi Zhang and Mofan Zhang and Ru Zhang and Wang Zhang and Hang Zhu and Jinhua Zhu and Jiaze Chen and Jiangjie Chen and Chengyi Wang and Hongli Yu and Yuxuan Song and Xiangpeng Wei and Hao Zhou and Jingjing Liu and Wei-Ying Ma and Ya-Qin Zhang and Lin Yan and Yonghui Wu and Mingxuan Wang},
    booktitle={The Thirty-ninth Annual Conference on Neural Information Processing Systems},
    year={2025},
    url={https://openreview.net/forum?id=2a36EMSSTp}
}

@inproceedings{
    mukherjee2025reinforcement,
    title={Reinforcement Learning Finetunes Small Subnetworks in Large Language Models},
    author={Sagnik Mukherjee and Lifan Yuan and Dilek Hakkani-T{\"u}r and Hao Peng},
    booktitle={The Thirty-ninth Annual Conference on Neural Information Processing Systems},
    year={2025},
    url={https://openreview.net/forum?id=0NdS4xCngO}
}

@article{liao2025beyond,
  title={Beyond Clicking: A Step Towards Generalist GUI Grounding via Text Dragging},
  author={Liao, Zeyi and Lu, Yadong and Gou, Boyu and Sun, Huan and Awadallah, Ahmed},
  journal={arXiv preprint arXiv:2601.06031},
  year={2025}
}

@misc{chen2025retainingdoingroleonpolicy,
      title={Retaining by Doing: The Role of On-Policy Data in Mitigating Forgetting}, 
      author={Howard Chen and Noam Razin and Karthik Narasimhan and Danqi Chen},
      year={2025},
      eprint={2510.18874},
      archivePrefix={arXiv},
      primaryClass={cs.LG},
      url={https://arxiv.org/abs/2510.18874}, 
}

@inproceedings{deng2023mind2web,
     author = {Deng, Xiang and Gu, Yu and Zheng, Boyuan and Chen, Shijie and Stevens, Sam and Wang, Boshi and Sun, Huan and Su, Yu},
     booktitle = {Advances in Neural Information Processing Systems},
     editor = {A. Oh and T. Naumann and A. Globerson and K. Saenko and M. Hardt and S. Levine},
     pages = {28091--28114},
     publisher = {Curran Associates, Inc.},
     title = {Mind2Web: Towards a Generalist Agent for the Web},
     url = {https://proceedings.neurips.cc/paper_files/paper/2023/file/5950bf290a1570ea401bf98882128160-Paper-Datasets_and_Benchmarks.pdf},
     volume = {36},
     year = {2023}
}

@inproceedings{
    zhou2024webarena,
    title={WebArena: A Realistic Web Environment for Building Autonomous Agents},
    author={Shuyan Zhou and Frank F. Xu and Hao Zhu and Xuhui Zhou and Robert Lo and Abishek Sridhar and Xianyi Cheng and Tianyue Ou and Yonatan Bisk and Daniel Fried and Uri Alon and Graham Neubig},
    booktitle={The Twelfth International Conference on Learning Representations},
    year={2024},
    url={https://openreview.net/forum?id=oKn9c6ytLx}
}

@inproceedings{seeact,
  author       = {Boyuan Zheng and
                  Boyu Gou and
                  Jihyung Kil and
                  Huan Sun and
                  Yu Su},
  title        = {GPT-4V(ision) is a Generalist Web Agent, if Grounded},
  booktitle    = {Forty-first International Conference on Machine Learning, {ICML} 2024,
                  Vienna, Austria, July 21-27, 2024},
  publisher    = {OpenReview.net},
  year         = {2024},
  url          = {https://openreview.net/forum?id=piecKJ2DlB},
  bibsource    = {dblp computer science bibliography, https://dblp.org}
}

@inproceedings{
  xue2025an,
  title={An Illusion of Progress? Assessing the Current State of Web Agents},
  author={Tianci Xue and Weijian Qi and Tianneng Shi and Chan Hee Song and Boyu Gou and Dawn Song and Huan Sun and Yu Su},
  booktitle={Second Conference on Language Modeling},
  year={2025},
  url={https://openreview.net/forum?id=6jZi4HSs6o}
}

@article{xie2024osworld,
  title={Osworld: Benchmarking multimodal agents for open-ended tasks in real computer environments},
  author={Xie, Tianbao and Zhang, Danyang and Chen, Jixuan and Li, Xiaochuan and Zhao, Siheng and Cao, Ruisheng and Toh, Jing Hua and Cheng, Zhoujun and Shin, Dongchan and Lei, Fangyu and others},
  journal={Advances in Neural Information Processing Systems},
  volume={37},
  pages={52040--52094},
  year={2024}
}

@inproceedings{wu2024copilot,
  title={OS-Copilot: Towards Generalist Computer Agents with Self-Improvement},
  author={Wu, Zhiyong and Han, Chengcheng and Ding, Zichen and Weng, Zhenmin and Liu, Zhoumianze and Yao, Shunyu and Yu, Tao and Kong, Lingpeng},
  booktitle={ICLR Workshop on Large Language Model (LLM) Agents},
year={2024}
}

@article{anthropic2024computeruse,
  title={Developing a computer use model},
  author={{Anthropic}},
  journal={Anthropic News},
  year={2024},
  month={Oct},
  url={https://www.anthropic.com/news/developing-computer-use}
}

@misc{wang2025uitars2technicalreportadvancing,
      title={UI-TARS-2 Technical Report: Advancing GUI Agent with Multi-Turn Reinforcement Learning}, 
      author={Haoming Wang and Haoyang Zou and Huatong Song and Jiazhan Feng and Junjie Fang and Junting Lu and Longxiang Liu and Qinyu Luo and Shihao Liang and Shijue Huang and Wanjun Zhong and Yining Ye and Yujia Qin and Yuwen Xiong and Yuxin Song and Zhiyong Wu and Aoyan Li and Bo Li and Chen Dun and Chong Liu and Daoguang Zan and Fuxing Leng and Hanbin Wang and Hao Yu and Haobin Chen and Hongyi Guo and Jing Su and Jingjia Huang and Kai Shen and Kaiyu Shi and Lin Yan and Peiyao Zhao and Pengfei Liu and Qinghao Ye and Renjie Zheng and Shulin Xin and Wayne Xin Zhao and Wen Heng and Wenhao Huang and Wenqian Wang and Xiaobo Qin and Yi Lin and Youbin Wu and Zehui Chen and Zihao Wang and Baoquan Zhong and Xinchun Zhang and Xujing Li and Yuanfan Li and Zhongkai Zhao and Chengquan Jiang and Faming Wu and Haotian Zhou and Jinlin Pang and Li Han and Qi Liu and Qianli Ma and Siyao Liu and Songhua Cai and Wenqi Fu and Xin Liu and Yaohui Wang and Zhi Zhang and Bo Zhou and Guoliang Li and Jiajun Shi and Jiale Yang and Jie Tang and Li Li and Qihua Han and Taoran Lu and Woyu Lin and Xiaokang Tong and Xinyao Li and Yichi Zhang and Yu Miao and Zhengxuan Jiang and Zili Li and Ziyuan Zhao and Chenxin Li and Dehua Ma and Feng Lin and Ge Zhang and Haihua Yang and Hangyu Guo and Hongda Zhu and Jiaheng Liu and Junda Du and Kai Cai and Kuanye Li and Lichen Yuan and Meilan Han and Minchao Wang and Shuyue Guo and Tianhao Cheng and Xiaobo Ma and Xiaojun Xiao and Xiaolong Huang and Xinjie Chen and Yidi Du and Yilin Chen and Yiwen Wang and Zhaojian Li and Zhenzhu Yang and Zhiyuan Zeng and Chaolin Jin and Chen Li and Hao Chen and Haoli Chen and Jian Chen and Qinghao Zhao and Guang Shi},
      year={2025},
      eprint={2509.02544},
      archivePrefix={arXiv},
      primaryClass={cs.AI},
      url={https://arxiv.org/abs/2509.02544}, 
}

@inproceedings{
    pan2024webcanvas,
    title={WebCanvas: Benchmarking Web Agents in Online Environments},
    author={Yichen Pan and Dehan Kong and Sida Zhou and Cheng Cui and Yifei Leng and Bing Jiang and Hangyu Liu and Yanyi Shang and Shuyan Zhou and Tongshuang Wu and Zhengyang Wu},
    booktitle={Agentic Markets Workshop at ICML 2024},
    year={2024},
    url={https://openreview.net/forum?id=O1FaGasJob}
}

@inproceedings{
    mirzadeh2025gsmsymbolic,
    title={{GSM}-Symbolic: Understanding the Limitations of Mathematical Reasoning in Large Language Models},
    author={Seyed Iman Mirzadeh and Keivan Alizadeh and Hooman Shahrokhi and Oncel Tuzel and Samy Bengio and Mehrdad Farajtabar},
    booktitle={The Thirteenth International Conference on Learning Representations},
    year={2025},
    url={https://openreview.net/forum?id=AjXkRZIvjB}
}

@article{anthropic2025computeruse,
  title={Claude 3.7 Sonnet and Claude Code},
  author={{Anthropic}},
  journal={Anthropic News},
  year={2025},
  month={Feb},
  url={https://www.anthropic.com/news/claude-3-7-sonnet}
}

@article{qin2025ui,
  title={Ui-tars: Pioneering automated gui interaction with native agents},
  author={Qin, Yujia and Ye, Yining and Fang, Junjie and Wang, Haoming and Liang, Shihao and Tian, Shizuo and Zhang, Junda and Li, Jiahao and Li, Yunxin and Huang, Shijue and others},
  journal={arXiv preprint arXiv:2501.12326},
  year={2025}
}

@misc{wang2025opencuaopenfoundationscomputeruse,
      title={OpenCUA: Open Foundations for Computer-Use Agents}, 
      author={Xinyuan Wang and Bowen Wang and Dunjie Lu and Junlin Yang and Tianbao Xie and Junli Wang and Jiaqi Deng and Xiaole Guo and Yiheng Xu and Chen Henry Wu and Zhennan Shen and Zhuokai Li and Ryan Li and Xiaochuan Li and Junda Chen and Boyuan Zheng and Peihang Li and Fangyu Lei and Ruisheng Cao and Yeqiao Fu and Dongchan Shin and Martin Shin and Jiarui Hu and Yuyan Wang and Jixuan Chen and Yuxiao Ye and Danyang Zhang and Dikang Du and Hao Hu and Huarong Chen and Zaida Zhou and Haotian Yao and Ziwei Chen and Qizheng Gu and Yipu Wang and Heng Wang and Diyi Yang and Victor Zhong and Flood Sung and Y. Charles and Zhilin Yang and Tao Yu},
      year={2025},
      eprint={2508.09123},
      archivePrefix={arXiv},
      primaryClass={cs.AI},
      url={https://arxiv.org/abs/2508.09123}, 
}

@misc{bai2025qwen3vltechnicalreport,
      title={Qwen3-VL Technical Report}, 
      author={Shuai Bai and Yuxuan Cai and Ruizhe Chen and Keqin Chen and Xionghui Chen and Zesen Cheng and Lianghao Deng and Wei Ding and Chang Gao and Chunjiang Ge and Wenbin Ge and Zhifang Guo and Qidong Huang and Jie Huang and Fei Huang and Binyuan Hui and Shutong Jiang and Zhaohai Li and Mingsheng Li and Mei Li and Kaixin Li and Zicheng Lin and Junyang Lin and Xuejing Liu and Jiawei Liu and Chenglong Liu and Yang Liu and Dayiheng Liu and Shixuan Liu and Dunjie Lu and Ruilin Luo and Chenxu Lv and Rui Men and Lingchen Meng and Xuancheng Ren and Xingzhang Ren and Sibo Song and Yuchong Sun and Jun Tang and Jianhong Tu and Jianqiang Wan and Peng Wang and Pengfei Wang and Qiuyue Wang and Yuxuan Wang and Tianbao Xie and Yiheng Xu and Haiyang Xu and Jin Xu and Zhibo Yang and Mingkun Yang and Jianxin Yang and An Yang and Bowen Yu and Fei Zhang and Hang Zhang and Xi Zhang and Bo Zheng and Humen Zhong and Jingren Zhou and Fan Zhou and Jing Zhou and Yuanzhi Zhu and Ke Zhu},
      year={2025},
      eprint={2511.21631},
      archivePrefix={arXiv},
      primaryClass={cs.CV},
      url={https://arxiv.org/abs/2511.21631}, 
}

@misc{sun2025scienceboardevaluatingmultimodalautonomous,
      title={ScienceBoard: Evaluating Multimodal Autonomous Agents in Realistic Scientific Workflows}, 
      author={Qiushi Sun and Zhoumianze Liu and Chang Ma and Zichen Ding and Fangzhi Xu and Zhangyue Yin and Haiteng Zhao and Zhenyu Wu and Kanzhi Cheng and Zhaoyang Liu and Jianing Wang and Qintong Li and Xiangru Tang and Tianbao Xie and Xiachong Feng and Xiang Li and Ben Kao and Wenhai Wang and Biqing Qi and Lingpeng Kong and Zhiyong Wu},
      year={2025},
      eprint={2505.19897},
      archivePrefix={arXiv},
      primaryClass={cs.AI},
      url={https://arxiv.org/abs/2505.19897}, 
}

@article{silver2016mastering,
  title={Mastering the game of Go with deep neural networks and tree search},
  author={Silver, David and Huang, Aja and Maddison, Chris J and Guez, Arthur and Sifre, Laurent and Van Den Driessche, George and Schrittwieser, Julian and Antonoglou, Ioannis and Panneershelvam, Veda and Lanctot, Marc and others},
  journal={nature},
  volume={529},
  number={7587},
  pages={484--489},
  year={2016},
  publisher={Nature Publishing Group}
}

@article{silver2018general,
  title={A general reinforcement learning algorithm that masters chess, shogi, and Go through self-play},
  author={Silver, David and Hubert, Thomas and Schrittwieser, Julian and Antonoglou, Ioannis and Lai, Matthew and Guez, Arthur and Lanctot, Marc and Sifre, Laurent and Kumaran, Dharshan and Graepel, Thore and others},
  journal={Science},
  volume={362},
  number={6419},
  pages={1140--1144},
  year={2018},
  publisher={American Association for the Advancement of Science}
}

@article{fang2025serl,
  title={SeRL: Self-Play Reinforcement Learning for Large Language Models with Limited Data},
  author={Fang, Wenkai and Liu, Shunyu and Zhou, Yang and Zhang, Kongcheng and Zheng, Tongya and Chen, Kaixuan and Song, Mingli and Tao, Dacheng},
  journal={arXiv preprint arXiv:2505.20347},
  year={2025}
}

@misc{liu2025spiralselfplayzerosumgames,
      title={SPIRAL: Self-Play on Zero-Sum Games Incentivizes Reasoning via Multi-Agent Multi-Turn Reinforcement Learning}, 
      author={Bo Liu and Leon Guertler and Simon Yu and Zichen Liu and Penghui Qi and Daniel Balcells and Mickel Liu and Cheston Tan and Weiyan Shi and Min Lin and Wee Sun Lee and Natasha Jaques},
      year={2025},
      eprint={2506.24119},
      archivePrefix={arXiv},
      primaryClass={cs.AI},
      url={https://arxiv.org/abs/2506.24119}, 
}

@article{wei2025toward,
  title={Toward Training Superintelligent Software Agents through Self-Play SWE-RL},
  author={Wei, Yuxiang and Sun, Zhiqing and McMilin, Emily and Gehring, Jonas and Zhang, David and Synnaeve, Gabriel and Fried, Daniel and Zhang, Lingming and Wang, Sida},
  journal={arXiv preprint arXiv:2512.18552},
  year={2025}
}

@inproceedings{
    zhao2025absolute,
    title={Absolute Zero: Reinforced Self-play Reasoning with Zero Data},
    author={Andrew Zhao and Yiran Wu and Yang Yue and Tong Wu and Quentin Xu and Yang Yue and Matthieu Lin and Shenzhi Wang and Qingyun Wu and Zilong Zheng and Gao Huang},
    booktitle={The Thirty-ninth Annual Conference on Neural Information Processing Systems},
    year={2025},
    url={https://openreview.net/forum?id=neZSGqhxDa}
}

@article{huang2025r,
  title={R-zero: Self-evolving reasoning llm from zero data},
  author={Huang, Chengsong and Yu, Wenhao and Wang, Xiaoyang and Zhang, Hongming and Li, Zongxia and Li, Ruosen and Huang, Jiaxin and Mi, Haitao and Yu, Dong},
  journal={arXiv preprint arXiv:2508.05004},
  year={2025}
}

@article{kuba2025language,
  title={Language self-play for data-free training},
  author={Kuba, Jakub Grudzien and Gu, Mengting and Ma, Qi and Tian, Yuandong and Mohan, Vijai and Chen, Jason},
  journal={arXiv preprint arXiv:2509.07414},
  year={2025}
}

@article{schulman2017proximal,
  title={Proximal policy optimization algorithms},
  author={Schulman, John and Wolski, Filip and Dhariwal, Prafulla and Radford, Alec and Klimov, Oleg},
  journal={arXiv preprint arXiv:1707.06347},
  year={2017}
}

@article{ouyang2022training,
  title={Training language models to follow instructions with human feedback},
  author={Ouyang, Long and Wu, Jeffrey and Jiang, Xu and Almeida, Diogo and Wainwright, Carroll and Mishkin, Pamela and Zhang, Chong and Agarwal, Sandhini and Slama, Katarina and Ray, Alex and others},
  journal={Advances in neural information processing systems},
  volume={35},
  pages={27730--27744},
  year={2022}
}

@article{cobbe2021training,
  title={Training verifiers to solve math word problems},
  author={Cobbe, Karl and Kosaraju, Vineet and Bavarian, Mohammad and Chen, Mark and Jun, Heewoo and Kaiser, Lukasz and Plappert, Matthias and Tworek, Jerry and Hilton, Jacob and Nakano, Reiichiro and others},
  journal={arXiv preprint arXiv:2110.14168},
  year={2021}
}

@inproceedings{
    zheng2025act,
    title={Act Only When It Pays: Efficient Reinforcement Learning for {LLM} Reasoning via Selective Rollouts},
    author={Haizhong Zheng and Yang Zhou and Brian R. Bartoldson and Bhavya Kailkhura and Fan Lai and Jiawei Zhao and Beidi Chen},
    booktitle={ES-FoMo III: 3rd Workshop on Efficient Systems for Foundation Models},
    year={2025},
    url={https://openreview.net/forum?id=23W5YZHGh9}
}

@misc{Polaris2025,
    title = {POLARIS: A Post-Training Recipe for Scaling Reinforcement Learning on Advanced Reasoning Models},
    url = {https://hkunlp.github.io/blog/2025/Polaris},
    author = {An, Chenxin and Xie, Zhihui and Li, Xiaonan and Li, Lei and Zhang, Jun and Gong, Shansan and Zhong, Ming and Xu, Jingjing and Qiu, Xipeng and Wang, Mingxuan and Kong, Lingpeng},
    year = {2025}
}

@article{shao2024deepseekmath,
  title={Deepseekmath: Pushing the limits of mathematical reasoning in open language models},
  author={Shao, Zhihong and Wang, Peiyi and Zhu, Qihao and Xu, Runxin and Song, Junxiao and Bi, Xiao and Zhang, Haowei and Zhang, Mingchuan and Li, YK and Wu, Yang and others},
  journal={arXiv preprint arXiv:2402.03300},
  year={2024}
}

@article{lai2025computerrl,
  title={Computerrl: Scaling end-to-end online reinforcement learning for computer use agents},
  author={Lai, Hanyu and Liu, Xiao and Zhao, Yanxiao and Xu, Han and Zhang, Hanchen and Jing, Bohao and Ren, Yanyu and Yao, Shuntian and Dong, Yuxiao and Tang, Jie},
  journal={arXiv preprint arXiv:2508.14040},
  year={2025}
}

@article{openaigpt5,
  title={Introducing GPT-5},
  author={{OpenAI}},
  journal={OpenAI Blog},
  year={2025},
  month={Aug},
  url={https://openai.com/index/introducing-gpt-5/}
}

@article{wang2025nemotron,
  title={Nemotron-Cascade: Scaling Cascaded Reinforcement Learning for General-Purpose Reasoning Models},
  author={Wang, Boxin and Lee, Chankyu and Lee, Nayeon and Lin, Sheng-Chieh and Dai, Wenliang and Chen, Yang and Chen, Yangyi and Yang, Zhuolin and Liu, Zihan and Shoeybi, Mohammad and others},
  journal={arXiv preprint arXiv:2512.13607},
  year={2025}
}

@inproceedings{
    feng2025groupingroup,
    title={Group-in-Group Policy Optimization for {LLM} Agent Training},
    author={Lang Feng and Zhenghai Xue and Tingcong Liu and Bo An},
    booktitle={The Thirty-ninth Annual Conference on Neural Information Processing Systems},
    year={2025},
    url={https://openreview.net/forum?id=QXEhBMNrCW}
}

@article{sheng2024hybridflow,
  title   = {HybridFlow: A Flexible and Efficient RLHF Framework},
  author  = {Guangming Sheng and Chi Zhang and Zilingfeng Ye and Xibin Wu and Wang Zhang and Ru Zhang and Yanghua Peng and Haibin Lin and Chuan Wu},
  year    = {2024},
  journal = {arXiv preprint arXiv: 2409.19256}
}

@article{ladosz2022exploration,
  title={Exploration in deep reinforcement learning: A survey},
  author={Ladosz, Pawel and Weng, Lilian and Kim, Minwoo and Oh, Hyondong},
  journal={Information Fusion},
  volume={85},
  pages={1--22},
  year={2022},
  publisher={Elsevier}
}

@inproceedings{
    diaz-bone2025discover,
    title={{DISCOVER}: Automated Curricula for Sparse-Reward Reinforcement Learning},
    author={Leander Diaz-Bone and Marco Bagatella and Jonas H{\"u}botter and Andreas Krause},
    booktitle={The Exploration in AI Today Workshop at ICML 2025},
    year={2025},
    url={https://openreview.net/forum?id=0kjOajnJQM}
}

@inproceedings{paul-etal-2024-refiner,
    title = "{REFINER}: Reasoning Feedback on Intermediate Representations",
    author = "Paul, Debjit  and
      Ismayilzada, Mete  and
      Peyrard, Maxime  and
      Borges, Beatriz  and
      Bosselut, Antoine  and
      West, Robert  and
      Faltings, Boi",
    editor = "Graham, Yvette  and
      Purver, Matthew",
    booktitle = "Proceedings of the 18th Conference of the European Chapter of the Association for Computational Linguistics (Volume 1: Long Papers)",
    month = mar,
    year = "2024",
    address = "St. Julian{'}s, Malta",
    publisher = "Association for Computational Linguistics",
    url = "https://aclanthology.org/2024.eacl-long.67/",
    pages = "1100--1126"
}

@misc{yuan2023scalingrelationshiplearningmathematical,
      title={Scaling Relationship on Learning Mathematical Reasoning with Large Language Models}, 
      author={Zheng Yuan and Hongyi Yuan and Chengpeng Li and Guanting Dong and Keming Lu and Chuanqi Tan and Chang Zhou and Jingren Zhou},
      year={2023},
      eprint={2308.01825},
      archivePrefix={arXiv},
      primaryClass={cs.CL},
      url={https://arxiv.org/abs/2308.01825}, 
}

@inproceedings{shinn2023reflexion,
  title={Reflexion: Language agents with verbal reinforcement learning},
  author={Shinn, Noah and Cassano, Federico and Gopinath, Ashwin and Narasimhan, Karthik and Yao, Shunyu},
  booktitle={Advances in Neural Information Processing Systems},
  volume={36},
  pages={8634--8652},
  year={2023}
}

@misc{xue2023rcotdetectingrectifyingfactual,
      title={RCOT: Detecting and Rectifying Factual Inconsistency in Reasoning by Reversing Chain-of-Thought}, 
      author={Tianci Xue and Ziqi Wang and Zhenhailong Wang and Chi Han and Pengfei Yu and Heng Ji},
      year={2023},
      eprint={2305.11499},
      archivePrefix={arXiv},
      primaryClass={cs.CL},
      url={https://arxiv.org/abs/2305.11499}, 
}

@inproceedings{
    gou2025navigating,
    title={Navigating the Digital World as Humans Do: Universal Visual Grounding for {GUI} Agents},
    author={Boyu Gou and Ruohan Wang and Boyuan Zheng and Yanan Xie and Cheng Chang and Yiheng Shu and Huan Sun and Yu Su},
    booktitle={The Thirteenth International Conference on Learning Representations},
    year={2025},
    url={https://openreview.net/forum?id=kxnoqaisCT}
}

@article{Gu2025WebDreamer,
  author    = {Yu Gu and Kai Zhang and Yuting Ning and Boyuan Zheng and Boyu Gou and Tianci Xue and Cheng Chang and Sanjari Srivastava and Yanan Xie and Peng Qi and Huan Sun and Yu Su},
  title     = {Is Your LLM Secretly a World Model of the Internet? Model-Based Planning for Web Agents},
  journal   = {Transactions on Machine Learning Research},
  year      = {2025}
}

@article{wei2025webagent,
  title={Webagent-r1: Training web agents via end-to-end multi-turn reinforcement learning},
  author={Wei, Zhepei and Yao, Wenlin and Liu, Yao and Zhang, Weizhi and Lu, Qin and Qiu, Liang and Yu, Changlong and Xu, Puyang and Zhang, Chao and Yin, Bing and others},
  journal={arXiv preprint arXiv:2505.16421},
  year={2025}
}

@inproceedings{
    qi2025webrl,
    title={Web{RL}: Training {LLM} Web Agents via Self-Evolving Online Curriculum Reinforcement Learning},
    author={Zehan Qi and Xiao Liu and Iat Long Iong and Hanyu Lai and Xueqiao Sun and Jiadai Sun and Xinyue Yang and Yu Yang and Shuntian Yao and Wei Xu and Jie Tang and Yuxiao Dong},
    booktitle={The Thirteenth International Conference on Learning Representations},
    year={2025},
    url={https://openreview.net/forum?id=oVKEAFjEqv}
}

@article{liu2025scalecua,
  title={Scalecua: Scaling open-source computer use agents with cross-platform data},
  author={Liu, Zhaoyang and Xie, JingJing and Ding, Zichen and Li, Zehao and Yang, Bowen and Wu, Zhenyu and Wang, Xuehui and Sun, Qiushi and Liu, Shi and Wang, Weiyun and others},
  journal={arXiv preprint arXiv:2509.15221},
  year={2025}
}

@article{feizi2025grounding,
  title={Grounding Computer Use Agents on Human Demonstrations},
  author={Feizi, Aarash and Nayak, Shravan and Jian, Xiangru and Lin, Kevin Qinghong and Li, Kaixin and Awal, Rabiul and L{\`u}, Xing Han and Obando-Ceron, Johan and Rodriguez, Juan A and Chapados, Nicolas and others},
  journal={arXiv preprint arXiv:2511.07332},
  year={2025}
}

@article{sun2025seagent,
  title={Seagent: Self-evolving computer use agent with autonomous learning from experience},
  author={Sun, Zeyi and Liu, Ziyu and Zang, Yuhang and Cao, Yuhang and Dong, Xiaoyi and Wu, Tong and Lin, Dahua and Wang, Jiaqi},
  journal={arXiv preprint arXiv:2508.04700},
  year={2025}
}

@article{he2025efficient,
  title={Efficient Agent Training for Computer Use},
  author={He, Yanheng and Jin, Jiahe and Liu, Pengfei},
  journal={arXiv preprint arXiv:2505.13909},
  year={2025}
}

@article{yang2025ultracua,
  title={UltraCUA: A Foundation Model for Computer Use Agents with Hybrid Action},
  author={Yang, Yuhao and Yang, Zhen and Dou, Zi-Yi and Nguyen, Anh and You, Keen and Attia, Omar and Szot, Andrew and Feng, Michael and Ramrakhya, Ram and Toshev, Alexander and others},
  journal={arXiv preprint arXiv:2510.17790},
  year={2025}
}

@article{awadallah2025fara,
  title={Fara-7B: An Efficient Agentic Model for Computer Use},
  author={Awadallah, Ahmed and Lara, Yash and Magazine, Raghav and Mozannar, Hussein and Nambi, Akshay and Pandya, Yash and Rajeswaran, Aravind and Rosset, Corby and Taymanov, Alexey and Vineet, Vibhav and others},
  journal={arXiv preprint arXiv:2511.19663},
  year={2025}
}

@article{cuvin2025decepticon,
  title={DECEPTICON: How Dark Patterns Manipulate Web Agents},
  author={Cuvin, Phil and Zhu, Hao and Yang, Diyi},
  journal={arXiv preprint arXiv:2512.22894},
  year={2025}
}

@article{schrittwieser2020mastering,
  title={Mastering atari, go, chess and shogi by planning with a learned model},
  author={Schrittwieser, Julian and Antonoglou, Ioannis and Hubert, Thomas and Simonyan, Karen and Sifre, Laurent and Schmitt, Simon and Guez, Arthur and Lockhart, Edward and Hassabis, Demis and Graepel, Thore and others},
  journal={Nature},
  volume={588},
  number={7839},
  pages={604--609},
  year={2020},
  publisher={Nature Publishing Group UK London}
}

@article{liu2025agent0,
  title={Agent0-VL: Exploring Self-Evolving Agent for Tool-Integrated Vision-Language Reasoning},
  author={Liu, Jiaqi and Xiong, Kaiwen and Xia, Peng and Zhou, Yiyang and Ji, Haonian and Feng, Lu and Han, Siwei and Ding, Mingyu and Yao, Huaxiu},
  journal={arXiv preprint arXiv:2511.19900},
  year={2025}
}

@article{bai2022training,
  title={Training a helpful and harmless assistant with reinforcement learning from human feedback},
  author={Bai, Yuntao and Jones, Andy and Ndousse, Kamal and Askell, Amanda and Chen, Anna and DasSarma, Nova and Drain, Dawn and Fort, Stanislav and Ganguli, Deep and Henighan, Tom and others},
  journal={arXiv preprint arXiv:2204.05862},
  year={2022}
}

@article{bai2022constitutional,
  title={Constitutional ai: Harmlessness from ai feedback},
  author={Bai, Yuntao and Kadavath, Saurav and Kundu, Sandipan and Askell, Amanda and Kernion, Jackson and Jones, Andy and Chen, Anna and Goldie, Anna and Mirhoseini, Azalia and McKinnon, Cameron and others},
  journal={arXiv preprint arXiv:2212.08073},
  year={2022}
}

@misc{kyidlicecek2025mathverify,
  title={Math-Verify: Math Verification Library},
  author={Kydlíček, Hynek},
  year={2025},
  version={0.6.1},
  url={https://github.com/huggingface/Math-Verify},
  license={Apache-2.0}
}

@article{wei2025swe,
  title={Swe-rl: Advancing llm reasoning via reinforcement learning on open software evolution},
  author={Wei, Yuxiang and Duchenne, Olivier and Copet, Jade and Carbonneaux, Quentin and Zhang, Lingming and Fried, Daniel and Synnaeve, Gabriel and Singh, Rishabh and Wang, Sida I},
  journal={arXiv preprint arXiv:2502.18449},
  year={2025}
}

@article{zhang2025agent,
  title={Agent learning via early experience},
  author={Zhang, Kai and Chen, Xiangchao and Liu, Bo and Xue, Tianci and Liao, Zeyi and Liu, Zhihan and Wang, Xiyao and Ning, Yuting and Chen, Zhaorun and Fu, Xiaohan and others},
  journal={arXiv preprint arXiv:2510.08558},
  year={2025}
}

@misc{ui-tars-15-seed,
  title = {UI-TARS-1.5},
  author = {ByteDance Seed},
  howpublished = {\url{https://seed-tars.com/1.5}},
  year = {2025},
}

@inproceedings{
    guan2025rstarmath,
    title={rStar-Math: Small {LLM}s Can Master Math Reasoning with Self-Evolved Deep Thinking},
    author={Xinyu Guan and Li Lyna Zhang and Yifei Liu and Ning Shang and Youran Sun and Yi Zhu and Fan Yang and Mao Yang},
    booktitle={Forty-second International Conference on Machine Learning},
    year={2025},
    url={https://openreview.net/forum?id=5zwF1GizFa}
}

@misc{zhang2025interplaypretrainingmidtrainingrl,
      title={On the Interplay of Pre-Training, Mid-Training, and RL on Reasoning Language Models}, 
      author={Charlie Zhang and Graham Neubig and Xiang Yue},
      year={2025},
      eprint={2512.07783},
      archivePrefix={arXiv},
      primaryClass={cs.CL},
      url={https://arxiv.org/abs/2512.07783}, 
}

@inproceedings{
    lu2025agentrewardbench,
    title={AgentRewardBench: Evaluating Automatic Evaluations of Web Agent Trajectories},
    author={Xing Han L{\`u} and Amirhossein Kazemnejad and Nicholas Meade and Arkil Patel and Dongchan Shin and Alejandra Zambrano and Karolina Stanczak and Peter Shaw and Christopher Pal and Siva Reddy},
    booktitle={Second Conference on Language Modeling},
    year={2025},
    url={https://openreview.net/forum?id=fQcUZMPIvu}
}

@article{zheng2026lifelong,
  title={Lifelong learning of large language model based agents: A roadmap},
  author={Zheng, Junhao and Shi, Chengming and Cai, Xidi and Li, Qiuke and Zhang, Duzhen and Li, Chenxing and Yu, Dong and Ma, Qianli},
  journal={IEEE Transactions on Pattern Analysis and Machine Intelligence},
  year={2026},
  publisher={IEEE}
}

@MISC{Ohio_Supercomputer_Center1987-dl,
 title   = "Ohio Supercomputer Center",
 author  = "{Ohio Supercomputer Center}",
 publisher = "Ohio Supercomputer Center",
 year   = "1987",
 url    = "https://ror.org/01apna436"
}

@misc{bai2024digirltraininginthewilddevicecontrol,
      title={DigiRL: Training In-The-Wild Device-Control Agents with Autonomous Reinforcement Learning}, 
      author={Hao Bai and Yifei Zhou and Mert Cemri and Jiayi Pan and Alane Suhr and Sergey Levine and Aviral Kumar},
      year={2024},
      eprint={2406.11896},
      archivePrefix={arXiv},
      primaryClass={cs.LG},
      url={https://arxiv.org/abs/2406.11896}, 
}

@article{gonzalez2025unreasonable,
  title={The unreasonable effectiveness of scaling agents for computer use},
  author={Gonzalez-Pumariega, Gonzalo and Tu, Vincent and Lee, Chih-Lun and Yang, Jiachen and Li, Ang and Wang, Xin Eric},
  journal={arXiv preprint arXiv:2510.02250},
  year={2025}
}

@inproceedings{agasheagent,
    title={Agent S2: A Compositional Generalist-Specialist Framework for Computer Use Agents},
    author={Saaket Agashe and Kyle Wong and Vincent Tu and Jiachen Yang and Ang Li and Xin Eric Wang},
    booktitle={Second Conference on Language Modeling},
    year={2025},
    url={https://openreview.net/forum?id=zg5is4GJ3R}
}

@article{chen2025self,
  title={Self-evolving curriculum for llm reasoning},
  author={Chen, Xiaoyin and Lu, Jiarui and Kim, Minsu and Zhang, Dinghuai and Tang, Jian and Pich{\'e}, Alexandre and Gontier, Nicolas and Bengio, Yoshua and Kamalloo, Ehsan},
  journal={arXiv preprint arXiv:2505.14970},
  year={2025}
}

@article{ishmam2026timewarp,
  title={TimeWarp: Evaluating Web Agents by Revisiting the Past},
  author={Ishmam, Md Farhan and Marino, Kenneth},
  journal={arXiv preprint arXiv:2603.04949},
  year={2026}
}

@article{legg2008machine,
  title={Machine super intelligence},
  author={Legg, Shane},
  year={2008}
}

@article{kirkpatrick2017overcoming,
  title={Overcoming catastrophic forgetting in neural networks},
  author={Kirkpatrick, James and Pascanu, Razvan and Rabinowitz, Neil and Veness, Joel and Desjardins, Guillaume and Rusu, Andrei A and Milan, Kieran and Quan, John and Ramalho, Tiago and Grabska-Barwinska, Agnieszka and others},
  journal={Proceedings of the national academy of sciences},
  volume={114},
  number={13},
  pages={3521--3526},
  year={2017},
  publisher={National Academy of Sciences}
}

@inproceedings{mallya2018packnet,
  title={Packnet: Adding multiple tasks to a single network by iterative pruning},
  author={Mallya, Arun and Lazebnik, Svetlana},
  booktitle={Proceedings of the IEEE conference on Computer Vision and Pattern Recognition},
  pages={7765--7773},
  year={2018}
}

@article{rusu2016progressive,
  title={Progressive neural networks},
  author={Rusu, Andrei A and Rabinowitz, Neil C and Desjardins, Guillaume and Soyer, Hubert and Kirkpatrick, James and Kavukcuoglu, Koray and Pascanu, Razvan and Hadsell, Raia},
  journal={arXiv preprint arXiv:1606.04671},
  year={2016}
}

@article{liu2026continual,
  title={Continual GUI Agents},
  author={Liu, Ziwei and Kang, Borui and Yuan, Hangjie and Zhao, Zixiang and Li, Wei and Zhu, Yifan and Feng, Tao},
  journal={arXiv preprint arXiv:2601.20732},
  year={2026}
}
